%% file: main.tex
\documentclass{WileyMSP-template}

\usepackage{colortbl}   %
\usepackage{xcolor}     %
\usepackage{adjustbox} 
\usepackage{graphicx}
\usepackage[numbers, sort, compress]{natbib}
\usepackage{steinmetz}
\graphicspath{{./images/}}
\usepackage[ruled,linesnumbered]{algorithm2e}
\usepackage{amsmath}
\usepackage{tabularx}
\usepackage{multirow}
\usepackage[table,xcdraw]{xcolor} %
\usepackage{url}

\begin{document}
\pagestyle{fancy}
\rhead{\includegraphics[width=2.5cm]{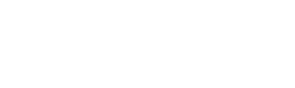}}

\title{Human-in-the-Loop Swarms: A Bionic Swarm Approach to Real-World Soil Mapping}

\maketitle

\author{Petras Swissler}
\author{Mohammadali Rashidioun}
\author{Nicholas Sahu}
\author{Raaid Kabir}
\author{Ayodeji Aderibigbe}
\author{Oladoyin Kolawole}

\dedication{}

\begin{affiliations}
Dr. Petras Swissler, Mr. Mohammadali Rashidioun\\
Department of Mechanical and Industrial Engineering, New Jersey Institute of Technology, Newark, New Jersey, United States of America\\
Email Address: petras.swissler@njit.edu

Mr. Nicholas Sahu\\
Department of Data Science, New Jersey Institute of Technology, Newark, New Jersey, United States of America

Mr. Raaid Kabir\\
University of Washington, Seattle, Washington, United States of America

Dr. Oladoyin Kolawole, Mr. Ayodeji Aderibigbe\\
Department of Civil and Environmental Engineering, New Jersey Institute of Technology, Newark, New Jersey, United States of America\\
Email Address: oladoyin.kolawole@njit.edu

\end{affiliations}


\begin{abstract}
\input{abstract}
\end{abstract}

\input{_overall_paper}

\medskip
\textbf{Acknowledgements} \par %
We would like to acknowledge the efforts of Saksha Poojari in assisting the connection of the sensors to the ESP32, as well as the assistance of those who assisted with performing the field tests: Myungjin Jung, Roberto Torres, Negar Namdar, and Leila Donyaparast.

This work is partially supported by a U.S. National Science Foundation Accelerating Research Translation cooperative agreement (TIP-2331429) and the NJIT Center for Translational Research. The opinions, findings, and conclusions, or recommendations expressed are those of the author(s) and do not necessarily reflect the views of the National Science Foundation.

\end{document}

%% file: abstract.tex
Swarm and field robotics face significant barriers to real-world validation due to the high cost and development time to deploy hardware. This paper introduces the ``Bionic Swarm,'' a novel system that lowers these barriers by abstracting away many of the tasks that are difficult to implement on robots but which do not contribute to the overall algorithm evaluation, giving these tasks to human users. These human users take directions from a smartphone web-app that takes measurements from Bluetooth-connected sensors and relays them to a centralized server. This server runs the swarm algorithm and directs actions to the human users. We evaluate this system through the experimental validation of a geotechnically-focused search algorithm named Score-Biased-Search, which functions by assigning a ``score'' to each location on a reconstructed map, then biases search patterns through areas of higher expected scores, and which exhibits superlinear map reconstruction relative to the number of search agents. After presenting simulation results for the algorithm, we then apply the algorithm on the Bionic Swarm platform to validate its function in a real-world, outdoor setting. This work demonstrates that this human-in-the-loop approach significantly lowers the barrier to entry for field and swarm robotics research.

%% file: _overall_paper.tex
\section{Introduction and Motivation}\label{sec1}
\input{introduction_and_motivation}

\section{Target Application for the Presented System}

\input{application_and_motivation}

\section{System Description}\label{sec2}
\input{system_overview}

\section{Simulated Algorithm Evaluation}
\input{simulation_results_new}

\section{Experimental System Evaluation}
\input{experimental_results}

\section{Discussion}
\input{discussion}

\section{Conclusions}\label{sec5}

\input{conclusions}

%% file: introduction_and_motivation.tex
oThe Mechanical Turk was a chess-playing automaton from the 18th century that amazed onlookers by its ability to play and prevail against many notable figures of its day.  The now-known secret to its function was that a human sat inside the automaton, puppeteering it from within. The reason for this is simple: at the time, it was much easier to implement the hardware aspect of a chess-playing system than the algorithmic and software aspects of such a system. Indeed, it took until 1912 to develop El Ajedrecista, the first actual artificial chess engine \cite{marsh2023chess}. This is the reverse of modern systems, where over a century of work has made the software and algorithmic aspects simple to implement, whereas the mechanical aspects (e.g. via a robot arm) would still be considered nontrivial to accomplish. 

This pattern extends to other types of robotic systems, especially swarm and field robots, where reliability and safety considerations can greatly complicate the necessary deployment and real-world validation of algorithms. In many cases, the ``interesting'' aspects of the algorithms such as processing the input of various sensors and determining the appropriate next course of action are not overly difficult to implement, but where actually executing those actions safely and reliably requires extensive work towards hardware development or the purchase of off-the-shelf hardware which can run from hundreds of dollars in the case of a system such as the Turtlebot3 \cite{TurtleBot3Overview} to upwards of a hundred thousand USD in the case of a system such as the Boston Dynamics Spot \cite{BostonDynamics2025}. Even when using off-the-shelf hardware, however, deploying robots in out-of-lab experiments requires extensive preparation to ensure that robots will operate safely and without damaging themselves in uncontrolled environments.

Economic pressures are particularly acute in swarm robotics, where the inherent use of large numbers of robots results in a high barrier to entry. It is also a field where custom hardware development is prevalent, often to address economic pressures \cite{rubenstein2014programmable, rezeck2023hero, jdeed2017spiderino}, and often to realize brand-new capabilities \cite{swissler2023fireantv3, liang2020freebot, saloutos2019spinbot, garnier2007alice, dorigo2013swarmanoid}. This bespoke hardware development can be extremely labor intensive, and can lead to significant compromises relative to what is realized in simulation. These compromises can take the form of reduced robot count relative to what would be required for the target application, reliance on remote control in the absence of autonomy, and the acceptance of suboptimal performance to avoid needing to re-design the robot. Worse, bespoke hardware development inevitably involves redundant work to realize aspects of the robot that are similar to existing systems (and are thus outside of the core research question) but which nonetheless require significant effort to get right; examples of such work includes drivetrain design, battery management, and other core robot features.

Despite these issues in field and swarm robotics, real-world testing is imperative, not only because any practical use robotics requires a physical manifestation, but also because real-world testing can reveal many interesting and important problems to be solved. Despite this, the need to deploy and sometimes to create robot hardware creates a high barrier to testing swarm concepts. This has the unfortunate effect of limiting the ability to test algorithms at earlier stages of their development, as well as making it likely that researchers will over-commit to certain design decisions (such as which sensors to use) in the absence of a way to test design features early on in the development process.

The premise explored in this paper is that \textbf{robot experiments should not inherently depend on robots}. Instead, we present a system that pairs the artificial sensory and processing aspects of a robotic system with the general applicability of a human to form a bionic swarm. The paper has two major contributions. First, we present Bionic Swarm, a generically-applicable system that enables the rapid testing of swarm algorithms in real-world environments. Second, we introduce a novel algorithm towards swarm geotechnical mapping, and evaluate it both in simulation and in real-world experiments using the Bionic Swarm platform.

The remainder of this paper is as follows. First, we describe a specific real-world application that would benefit from a swarm approach: soil pollution mapping. Next, we present the bionic system, including the algorithm to enable efficient geotechnical mapping. We then present simulated results for this algorithm and show the value of a swarm approach. This is followed by the results of a real-world experiment, as well as a discussion comparing the two sets of results, as well as comparing obstacles to deployment for a robot-based system versus the bionic swarm presented in this paper. Finally, we discuss future directions for this line of research in the conclusion.

%% file: application_and_motivation.tex
Geotechnical mapping, a cornerstone of modern engineering, involves the comprehensive analysis and visualization of underground (geologic) materials (soil, rock, and other materials) to inform construction, mining, and environmental projects in geotechnical engineering \citep{arshid2020regional, pan2021efficient, khan2022review, ijaz2021spatial, ijaz2023development, mamdooh2023gis, miao2024interpolation}. The advent of spatial and geostatistical analysis has revolutionized this field, providing sophisticated tools to model and predict geotechnical properties over large areas with greater accuracy \citep{wang2016combination, khalid2021application}. Spatial analysis, using Geographic Information Systems (GIS), enables the integration of various data types and sources, offering a spatial perspective that is crucial for understanding geotechnical variations across different terrains and conditions \citep{khan2022review, mamdooh2023gis, chamine2022role}. Meanwhile, geostatistical analysis offers rigorous methods for quantifying spatial relationships and patterns among geotechnical data, facilitating more precise estimations of subsurface conditions and associated uncertainties \citep{hassan2022geospatial, hassan2023statistical}. Together, these technologies enhance our ability to conduct risk assessments, design foundations, and manage natural resources, making geotechnical mapping an indispensable component of modern engineering endeavors. By leveraging spatial data and statistical models, geotechnical and geological engineers can characterize, optimize designs and mitigate geohazard risks more effectively, showcasing a significant evolution from traditional methods that were often time-consuming and less reliable \citep{pan2021efficient, hassan2023statistical, zhang2012return}. Over the past few decades, various methods and models have been proposed as the most suitable for field-scale characterization and imaging of geotechnical properties and behaviors, which are crucial for safe and economical construction practices.

Spatial analysis is one of the most popular methods in geotechnical mapping and involves the use of Geographic Information Systems (GIS) and other spatial tools to visualize, manipulate, and analyze geospatial data \citep{khan2022review, mamdooh2023gis, chamine2022role}. This GIS approach allows engineers and scientists to create detailed maps that depict and predict variations in soil composition \citep{lindberg2011mapping, omer2022mapping}, rock properties \citep{prasad2008deciphering, shirzadi2012gis}, and potential geological hazard features \citep{shirzadi2012gis, xie2006geographical, sun2008development, zolfaghari2008gis} across a given area. Such geotechnical maps are invaluable in identifying potential hazards, planning construction projects, and managing natural resources effectively \citep{lindberg2011mapping, xie2006geographical, sun2008development}.
In the context of geotechnical mapping, geostatistical techniques like Kriging, variogram analysis, and stochastic simulation enable the estimation of geological variables at unsampled locations \citep{pan2021efficient, miao2024interpolation, wang2016combination, pokhrel2013kriging}. This geostatistical approach is particularly advantageous in geotechnical engineering, where direct observations can be expensive or infeasible. By modeling the spatial continuity and variability of soil and rock properties, geostatistical methods can enhance the accuracy of site-specific property and behavior predictions, thereby reducing the uncertainty inherent in engineering and environmental decision-making, but the available geostatistical methods have not yet been able to achieve this feat in unknown search domains.

The integration of spatial and geostatistical analysis into geotechnical mapping has led to the development of more sophisticated, data-driven approaches to understanding the Earth's subsurface and estimating the spatial distribution of geotechnical parameters \citep{ijaz2021spatial, ijaz2023development, wang2016combination, hassan2022geospatial, hassan2023statistical, madani2018spatial, caballero2022geotechnical}. The results from these coupled analyses have provided an improvement in the precision of geotechnical assessments, and but there is still a challenge in improving its efficiency in handling large datasets, which is increasingly common with the rise of remote sensing technologies and automated data collection methods \citep{arshid2020regional, ijaz2021spatial, ijaz2023development, de2012integrating}. As a result, the capacity to conduct thorough, accurate geotechnical evaluations over extensive areas has been inhibited, thereby limiting large-scale infrastructure projects and resource management initiatives in complex and challenging geologic domains \citep{nasipuri2006development, roslee2012intergration, mahmoudi2021optimisation}.

However, despite these advancements, the field of geotechnical mapping faces ongoing challenges. The variability of geological environments, coupled with the limitations of current techniques and methods, often leads to complexities in data interpretation and model validation. Moreover, the integration of new data sources, such as satellite imagery and unmanned aerial vehicle (UAV) surveys, into existing geotechnical frameworks demands continuous adaptation and improvement of analytical techniques. There is also a growing need for better data integration and an interdisciplinary approach to address these multi-faceted challenges presented by complex geotechnical environments. Therefore, there is a need to develop an efficient and robust mapping algorithm with spatial and geostatistical analysis that enhances the depth and accuracy of geotechnical assessments with greater potential for automated pattern recognition and predictive modeling. 

In robotics, mapping of unknown environments is a common area of study \citep{lluvia2021active, smith1986representation, durrant2006simultaneous, thrun2000real}, with great consideration being given to how exactly robots will navigate an environment during such a task. This navigation typically relies on one of a variety of path-planning algorithms such as \citep{stentz1995D*, lavalle1998RRT, dorigo1997antcolonyoptimization, li2012efficient}, with A* being one of the simplest to implement. Work on mapping in the robotics field has expanded to include mapping by multi-agent systems \citep{albani2017monitoring, iser2010antslam, kegeleirs2021swarm, pires2021cooperative, zhang2022multi}, which often excel in parallelizable tasks.  
In most robotic mapping problems, however, there is a reliance on long-range sensors such as LIDAR and cameras which classify points in a map as being either occupied or unoccupied (binary characterization). This strongly contrasts with the sort of mapping typically done in geotechnical fields, where measurements are often hyper-local and the goal is to map some continuous (non-binary) variable throughout some continuous search domain (e.g., the concentration of a pollutant in a field). A common method of approaching this sort of mapping problem is to apply the Kriging reconstruction method \citep{wang2016combination, krige1951statistical, pulido2020kriging} based on relatively sparse measurements. We believe that the future of geotechnical sampling lies in in-situ measurements that enable continuous sampling and adaptive search patterns.

%% file: system_overview.tex
\begin{figure}
    \centering
    \includegraphics[width=0.75\textwidth]%
    {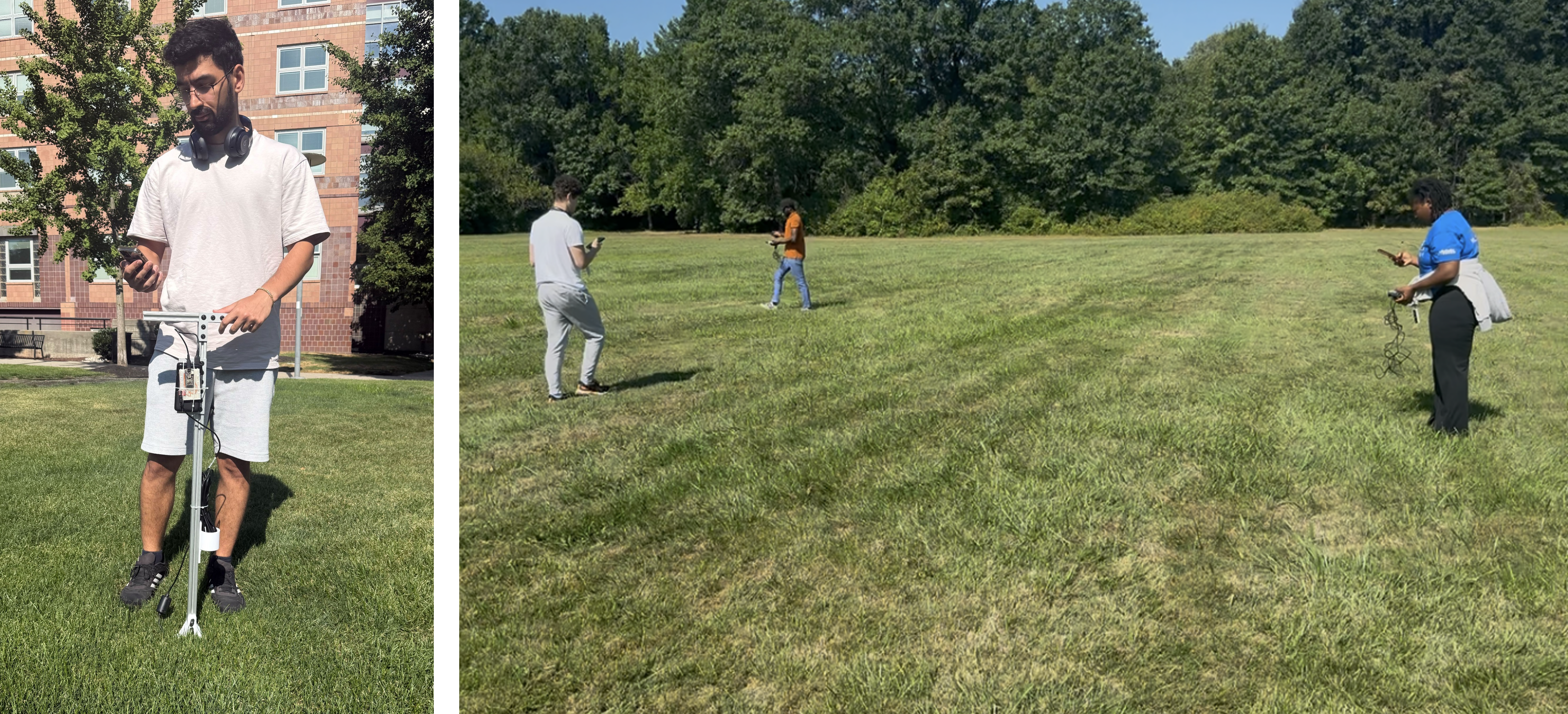}
    \caption{Users of  Bionic Swarm navigate an open field, guided via a smartphone and taking measurements using a soil sensor. Experiments were performed using a swarm of four human operators.}
    \label{fig:users}
\end{figure}

The presented system, shown in-use in Fig. \ref{fig:users}, consists of three general components, detailed in the subsections below. First, we present the backend architecture that enables the coordination between a centralized processing server and the distributed users in the field. We then present a search algorithm to enable intelligent mapping of the soil, which will be run on the centralized processing server. Next, we present the UI that users see on their smartphones when executing the search. Finally, we briefly present the soil sensors used in the experiments.

We name this system the ``Bionic Swarm'' since it augments human behavior via artificial sensing and super-human coordination.

\subsection{System overview and architecture}
\input{system_architecture}

\subsection{Search Algorithm}
\input{search_algorithm}

\subsection{User Interface}
\input{user_interface}

\subsection{Soil Sensors}
\input{soil_sensors}

%% file: system_architecture.tex
\begin{figure}
    \centering
    \includegraphics[width=0.5\textwidth]%
    {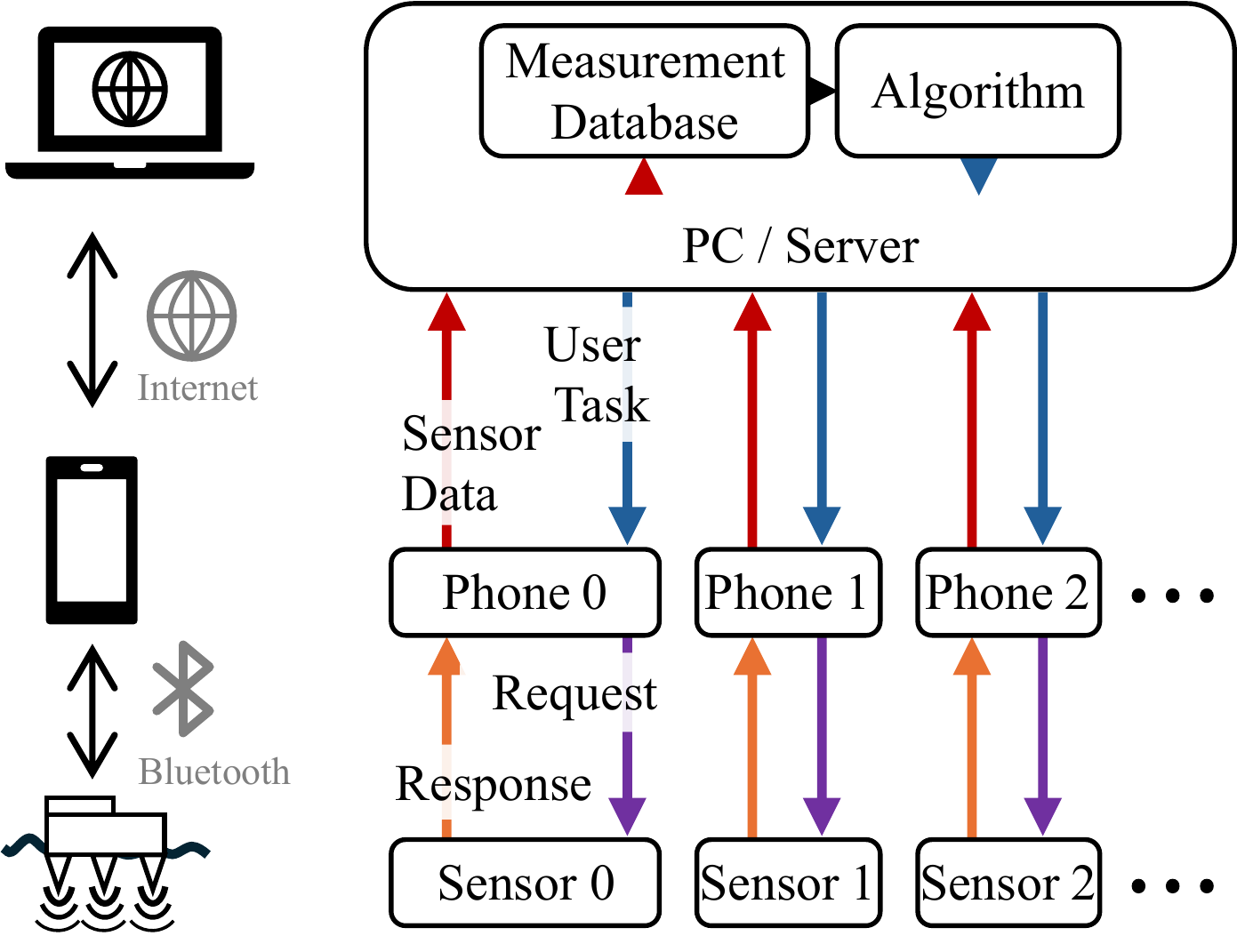}
    \caption{Sensors report data to smartphones, which then relay this data to a server. The server then provides the user a task to complete.}
    \label{fig:architecture}
\end{figure}

The Bionic Swarm, whose architecture is illustrated in Fig. \ref{fig:architecture}, offers a generic method for integrating mobile, real-world sensors into an algorithm running on a server. In the realization of this system presented here, the server consisted of a generic laptop hosting a server at a static IP address. This server is able to collect measurement data and run a python script that executes the algorithm to be tested. This data is pulled from an arbitrarily-large number of user smartphones, each of which is running a web app that users can open by simply scanning a QR code on the laptop screen. The server also sends per-user data to each smartphone that is displayed to the user via a customizable GUI (see Section \ref{GUI_Section}). This web app uses Bluetooth to transmit serial data to and from a ESP32-S3 microcontroller. In the system presented in this paper, the microcontroller merely relays sensor measurements from the TEROS 12 soil sensor \cite{meter2025teros12} (see Section \ref{SENSOR_Section}), but could be configured to read practically any sort of sensor, to control one or more actuators, or even to interface with an entire robot. Note also that the web app is able to access most phone sensors, including GPS and the IMU.

The source code for this system, including code for the server-run algorithm and the web app are available at \cite{code_folder}.

%% file: search_algorithm.tex
As part of the Bionic Swarm system, we present and validate a novel Score-Biased-Search (SBS) algorithm to efficiently map information-rich environments. The algorithm enables robotic agents to map a quantity of interest in a discretized 2D grid domain.

\subsubsection{Overview}
We assume that the search agents have the following capabilities:

\begin{enumerate}
    \item The ability to measure a quantity of interest at their current location, and that this measurement is fast relative to the speed of motion.
    \item The ability to move in any direction.
    \item The ability to know their position in the grid-world (for example, by using GPS) and to share this position.
\end{enumerate}

SBS consists of three major steps. First, agents take a measurement at their current position, and these measurements are then added to a shared measurement database containing measurement-location pairs for each of the search agents. Second, these measurement-location pairs are fed into an Ordinary Kriging reconstruction model to obtain the estimated value and uncertainty estimate at each cell of the discretized search domain. Note that because measurements are taken as the agents are moving, these measurements are inherently clustered (they are taken one after another while the agent is en route to the goal location), necessitating a covariance-aware method such as Kriging, rather than computationally cheaper methods such as Inverse Distance Weighting. Another benefit to using Kriging reconstruction is that it quantifies uncertainty, which will be useful for guiding exploration.

The final step of the algorithm involves the selection of a goal location and the routing of the agent to that goal location. A ``score map'' is computed for each agent using the Kriging-derived map (see discussion of algorithm parameters below), with the domain being divided among the agents using Voronoi partitioning \cite{aurenhammer1991voronoi} . The maximum computed score becomes the agent's goal destination. The route to this goal location is computed using A*, where the cost of entering a cell is equal to the inverse of its score, thus biasing agent movement through areas of highest expected score. Agents then take one step along the calculated optimal path before this process repeats.

\begin{algorithm}
    \caption{Overview of SBS}
    \label{alg_search}
    
    \BlankLine
    \textbf{load:} parameters \;
    \textbf{init:} meas $\leftarrow$ null \;
    \textbf{init:} agent\_positions $\leftarrow$ initial\_positions \;

    \For{$i = 1 \dots \text{num\_steps}$}{
        \tcp{All agents sample at their current location}
        \For{$r = 1 \dots \text{num\_agents}$}{
            meas.append(sample(agent(r))) \;
        }

        \tcp{Estimate the map using all measurements.}
        [estimates, uncertainties] $\leftarrow$ Kriging(meas) \;

        \tcp{Agents use the reconstructed map to determine the optimal goal location and the optimal route to that goal location.}
        \For{$r = 1 \dots \text{num\_agents}$}{
            map\_scores $\leftarrow$ ComputeScore(estimate, uncertainties, agent\_goal\_locations(r), agent\_locations, agent\_search\_parameters, weight\_prefer\_current\_goal) \;
            
            agent\_goal\_locations(r) $\leftarrow$ FindMax(map\_scores) \;
            
            optimal\_route $\leftarrow$ AStar(agent\_locations(r), agent\_goal\_locations(r), $\text{normalize(map\_scores)}^{-1}$ / weight\_step\_cost) \;

            agent\_positions(r) $\leftarrow$ Take step on optimal\_route \;
        }
    }
\end{algorithm}

\begin{figure}[]
    \centering
    \includegraphics[width=0.75\textwidth]%
    {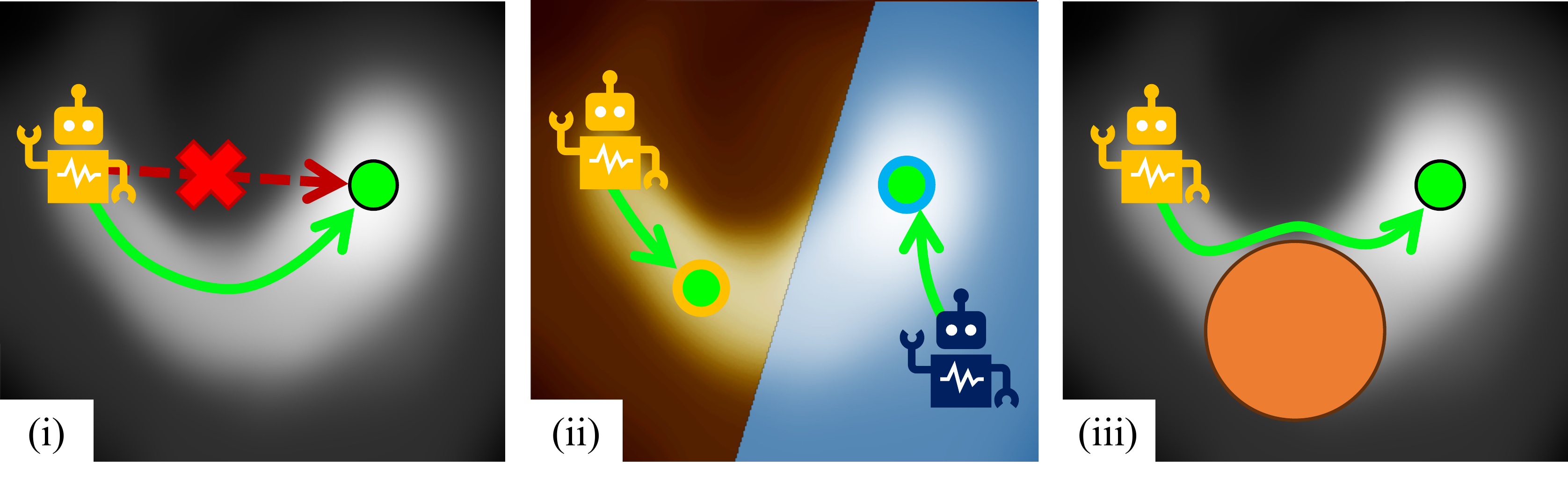}
    \caption{Illustration of three major aspects of search agent behavior. The background represent the evaluated score, where dark represents areas of low score and light represents areas of high score.}
    \label{fig:algorithm_overview}
\end{figure}

This algorithm manifests three general behaviors, illustrated in Fig. \ref{fig:algorithm_overview}.  First, instead of moving directly to the goal location (shown as a green circle), agents bias their movement towards regions with high score, often resulting in movement through areas of higher uncertainty. Partitioning the domain among multiple agents is accomplished by simply modifying the per-agent scores to implement Voronoi partitioning and  ensure full map coverage without overlap. Finally, because routing is accomplished via a standard path-planning technique, obstacles are easily accounted for in the search pattern.

\subsubsection{Parameter Selection}
Algorithm performance is largely dictated by the way in which map scores are calculated. These scores dictate not only the goal locations chosen but also the paths that agents take to these goal locations. Through iterative tests, we identified several parameters that appeared to be important in the performance of this algorithm. The overall score is calculated as the sum of these weights multiplied by normalized versions of their relevant quantities; this normalization helps avoid degenerate conditions.

To determine the best set of parameters for the algorithm, we ran a full factorial test in which we tested combinations of  different orders of magnitude for each weight parameter [0, 0.1, 1, 10, and 100] with 100 replicants. To prevent biasing these parameters towards a given type of environment (e.g., an environment that is very ``rich'' or an environment that is very evenly distributed), we generated a set of synthetic maps using Fractional Brownian Fields , then randomly modifying how these fields were translated to map values.

\begin{figure}[]
    \centering
    \includegraphics[width=0.75\textwidth]%
    {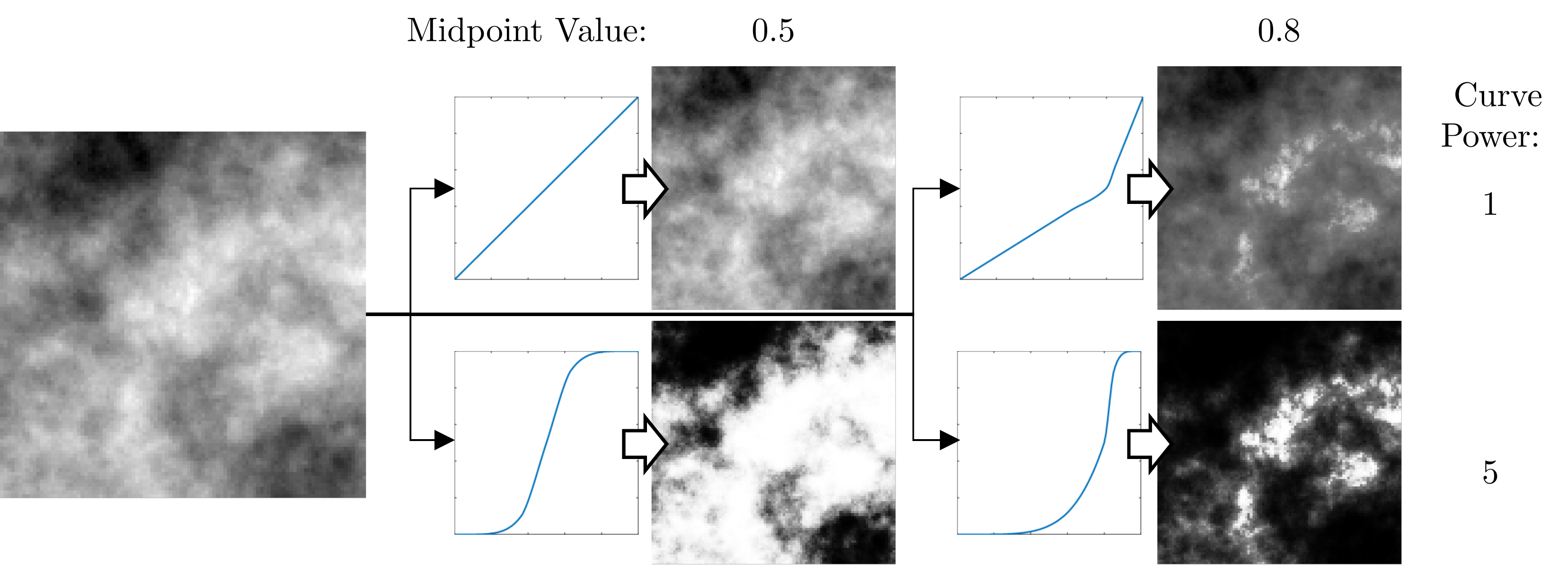}
    \caption{The parameters of the S-Curve change the distribution of the variable of interest in the domain, either by changing the overall availability of the variable (Midpoint Value), or by changing whether the distribution of the variable is evenly distributed or highly concentrated (Curve Power).}
    \label{fig:scurve_show}
\end{figure}

We chose to use Fractional Brownian Fields due to their common use to generate random terrain \citep{van1995dynamic},
with the specific implementation we used being adapted from \citep{Botev2024FractionalBrownian} using a Hurst parameter of 0.7.
These fields were then converted into a  100 $\times$ 100 pixel sections, with values being normalized between 0 and 1. For parameter selection runs only, we tune the expression of this map using an S-curve so map input values 0\textendash1 to output values 0\textendash1. The S-curve function we use takes in two parameters: $threshold\_value$ determines the ``richness'' of the environment by setting the input value that will have an output value of 0.5 (for high $threshold\_value$ this means that most input values will be mapped to values below 0.5 and vice versa); the parameter $curve\_power$ determines how smoothly or bimodally input values are mapped to output values (for high $curve\_power$, output values will tend to be close to either 0 or 1, with very few in between).
Fig. \ref{fig:scurve_show} illustrates the effects of these parameters. This approach covers a broad range of potential environment types and helps ensure that the selected parameters will be usable in real-world deployments.

We then ran tests using randomized $threshold\_values$ and $curve\_powers$, and then selected the combination of parameter values that gave the overall best results, with tests being performed using a single mapping agent. These parameters are as follows:

\begin{itemize}
    \item \verb|weight_expected_value| is multiplied by the expected value of a given location according to the Kriging reconstruction. This expected value is normalized to between 0 (smallest expected value in the domain) and 1 (highest expected value in the domain) This term biases search towards areas of high expected value (e.g., with a high pollution concentration). We found that a value of 1 for this weight gave the best performance.
    
    \item \verb|weight_uncertainty| is multiplied by the reported uncertainty of the Kriging reconstruction for a given location. This uncertainty is normalized between 0 (lowest uncertainty in the domain) and 1 (highest uncertainty in the domain). This term biases search towards under-explored regions. We found that a value of 10 for this weight gave the best overall performance.

    \item \verb|weight_prefer_center| is multiplied by the one minus the normalized distance from the center of the search domain; this normalized distance is a value between 0 (the center location) and 1 (the furthest point from center). This term biases the search towards the center of the search domain. We found that a value of 0.1 for this weight gave the best overall performance.
    
    \item \verb|weight_prefer_closeness| is multiplied by one minus the normalized distance of a given location from the agent's current location; this normalized distance is a value between 0 (closest point to the agent) and 1 (furthest point from the agent). This term biases search nearer the agent. We found that a value of 0.1 for this weight gave the best overall performance.

    \item \verb|weight_prefer_current_goal| is multiplied by one minus the normalized distance of a given location from the agent's current goal; this normalized distance is a value between 0 (closest point to the current goal) and 1 (furthest point from the current goal). This term helps eliminate instances where two disparate goal locations have similar scores, leading to the agent switching goals with every step, limiting exploration. We found that a value of 10 for this weight gave the best overall performance, although we eventually found that the best value for this weight varies based on the number of agents.
    
\end{itemize}

These order-of-magnitude results provide several key insights to what a search algorithm should consider. First, search algorithms should explore areas of high uncertainty. Second, there is benefit to maintaining the current goal, rather than switching to newly-discovered ``best'' goals. Finally, even though the relative worth of biasing the search to near the agent and near the center are less than the other quantities, the algorithm still performs better when these factors are considered.

In addition to these algorithm weights associated with evaluating the ``score'' of the points on the map, there is an additional algorithm parameter \verb|weight_step_cost| that describes the importance that the agents place on minimizing overall travel distance versus seeking to move through high-priority areas. Based on further experimentation we found that using a step cost of 0.01 yielded the best results, indicating that the best approach was to prioritize routing motion through ``good'' areas over simply minimizing travel distance, affirming the importance of intelligent routing through the search space.

%% file: user_interface.tex
\label{GUI_Section}

\begin{figure}
    \centering
    \includegraphics[height=3.5in]%
    {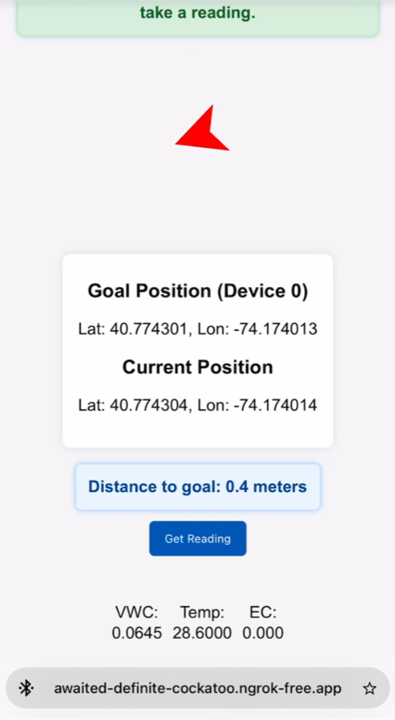}
    \caption{The interface used by the human to interact with the system. The red arrow points in the direction that the person should move, relative to their current orientation.}
    \label{fig:phone_app}
\end{figure}

The web app that serves as the user interface for the presented system is shown in Fig. \ref{fig:phone_app}. The red arrow at the top of the screen points towards the goal position and is the result of the app synthesizing the user's position (measured using the phone's GPS), the server-provided next measurement position, as well as the phone's compass. When the user is near the next measurement location, a green message appears at the top of the screen to prompt the user to take a measurement by touching the ``Get Reading'' button. Scrolling down reveals experiment tracking metrics, including the in-progress map.

This simple user interface is sufficient to enable human users to perform the basic actions necessary to enact the search algorithm while simultaneously allowing users the leeway to exercise their best judgment for things that may be difficult to anticipate and implement in a purely-robotic system, such as choosing an exact measurement position that avoids rocks, avoiding previously-unknown obstacles, etc.

The number of users able to connect to the system is bounded only by the the overall networking capacity of the server. In practice the limit is likely to be bounded by the availability of human operators than it is by networking considerations. In the experiments presented in Section \ref{Section_experiments}, we present experiments using four simultaneous users, though we did field-validate the ability to handle six simultaneous users using only networking that was available away from Wi-Fi connections.

%% file: soil_sensors.tex
\label{SENSOR_Section}

\begin{figure}[]
  \centering
  \includegraphics[width=0.5\textwidth]%
    {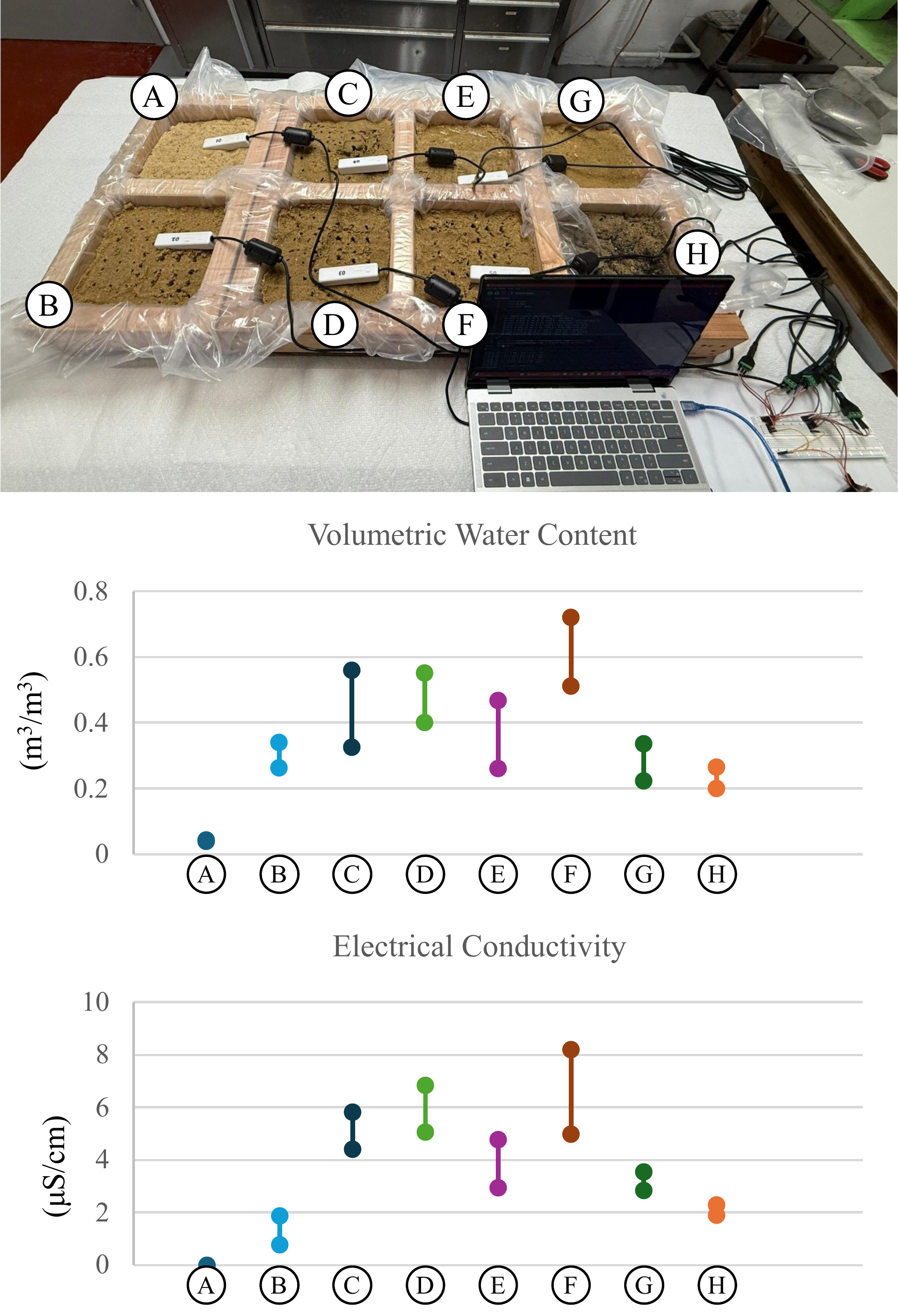}
  \caption{Experimental setup and results for soil samples: \textbf{A}--Dry soil (baseline), \textbf{B}--Freshwater-saturated soil, \textbf{C}--Saltwater-contaminated soil, \textbf{D}--Low pH soil, \textbf{E}--High pH soil, \textbf{F}--Chemically contaminated soil, \textbf{G}--Oil spill-contaminated soil, and \textbf{H}--Sewage-contaminated soil.}
  \label{fig:sensor_mapping}
\end{figure}

The soil sensors used in the presented system are the TEROS-12 sensors, which have the capability to sense both soil moisture and electrical conductivity. While the experiments in Section \ref{sec:experiments} merely mapped soil moisture, it is important to demonstrate that these sensors can be used to detect various types of pollution. To do this, we performed a characterization experiment in which various contaminants were added to eight 30 cm by 30 cm soil sample squares, and then the volumetric water content and the electrical conductivity were measured. 

The system demonstrated strong discriminatory capability across these samples. For instance, dry soil (Sample~A) exhibited the lowest VWC \((0.041~\mathrm{m}^3/\mathrm{m}^3)\) and zero EC, establishing a baseline for uncontaminated conditions. In contrast, saltwater-contaminated soil (Sample~C) produced the highest VWC \((0.616~\mathrm{m}^3/\mathrm{m}^3)\) and EC \((5.95~\mu\mathrm{S}/\mathrm{cm})\), while sewage-contaminated soil (Sample~H) recorded the highest EC \((6.60~\mu\mathrm{S}/\mathrm{cm})\), reflecting elevated ion concentrations. Similarly, soils with extreme pH values (Samples~D and E) and those contaminated with chemicals or oil spills (Samples~F and G) displayed distinct VWC and EC signatures, enabling accurate differentiation of contamination types. These results confirm the ability for the sensor to detect and characterize diverse contamination scenarios, a critical step toward real-world hazard mapping. Importantly, integrating VWC and EC with temperature measurements enhanced the system’s capacity to distinguish pollutants with high accuracy, even under overlapping conditions.

%% file: simulation_results_new.tex
\subsection{Evaluation Metrics}

To establish a baseline from which to examine system performance, as well as to evaluate the overall effectiveness of the proposed algorithm, we evaluated SBS in a series of simulated trials. This evaluation is centered around quantifying how quickly and accurately the algorithm characterizes the search domain. We place a special emphasis on the ability to identify ``high value'' regions e.g., areas with a high amount of pollution. 

In addition to standard Sum of Squared Error (SSE) metrics, we also evaluate the Critical Success Index (CSI). In the context of spatial mapping, we use CSI as a rigorous metric to quantify how well the reconstructed map captures specific characterization levels, defined as percentile thresholds of the given variable of interest. In this paper, we term this combination of CSI and thresholding as the ``Characterization Accuracy'' \textit{CAX}, where ``X'' denotes the percentile threshold being evaluated (e.g., CA50 when evaluating a percentile threshold of 50\%). It is calculated as follows:

\begin{equation}
    \begin{split}
    {\rm CAX} = \frac{{\rm TP_X}}{{\rm TP_X} + {\rm FP_X} + {\rm FN_X}}
    \end{split}
    \vspace{3mm}
\end{equation}

Where: TP is the count of true positives, where a cell is correctly identified as exceeding the threshold \textit{X} in both the reconstructed and ground-truth maps; FP is the count of false positives, where a cell that has a value above the threshold \textit{X} in the reconstructed map but does not in the ground-truth map; and FN is the count of false negatives, where a cell has a value below the threshold \textit{X} in the reconstructed map but in truth has a value above that threshold in the ground-truth map. In an ideal case where the ground-truth map was perfectly reconstructed (i.e., FP and FN are always zero), all characterization accuracies would be equal to 1.0, but this is unrealistic to achieve with purely local sensing short of an exhaustive search.

In practice, this metric becomes significantly more demanding at higher characterization levels due to increasing sensitivity to random noise. As a simple example, consider a completely random relationship between the reconstructed values and the ground truth values. When evaluating a threshold of 50\%, there would be an equal probability for each of TP, FP, and FN, giving $CA50_{random} = 1/3 \approx 0.33$. Calculating this for a threshold of 90\% would give $CA90_{random} = 1/19 \approx 0.05$. %

Interpretation of the presented results should be viewed through this lens of increasing difficulty for high characterization thresholds, especially considering the definitional randomness of a Fractional Brownian Field.
Additionally, we must consider the ``shape'' of the curve defined by the characterization accuracy relative to time. For example, a system that quickly improves before asymptotically approaching a given characterization accuracy is preferable to a system that linearly approaches that same characterization accuracy since mapping activities can be unexpectedly cut short by any number of external factors. Thus, early-stage advantages in map reconstruction accuracy make the overall system more operationally robust. In this paper we do not present any quantitative metrics of shape but instead rely on qualitative evaluations of the various graphs we present.

Note that these evaluation metrics are not necessarily monotonic: exploration of a new region may lead to shifting, incorrect assumptions about an unexplored region.

\subsection{Simulated Evaluation}

To evaluate performance, a new set of 100 maps are created of size 100 cells by 100 cells, with each of the results below being performed on the same set of maps. Evaluation metrics are considered in the context of cumulative search time: one agent searching for 800 steps is considered equivalent to two agents searching for 400 steps. This search duration corresponds to a search that samples 8\% of the map.

We first examine traditional approaches to enable direct benchmarking, starting with a na\"ive search, followed by analyses of single and multi-agent searches for a baseline point-to-point search compared to the proposed approach. We also examine the performance of the proposed algorithm in environments with obstacles. In each case, trials are run using unattenuated test maps (i.e., without using the S-Curve attenuation that was used for parameter selection).

Results are summarized in Table \ref{tab:full Results}, and are detailed below using two types of graphs. The first is a chart showing the mean CAX for several threshold values over time for the 100 tested maps. In a high-performing system, the values for CAX would increase quickly, though because of the random nature of the environment and the lack of long-range measurements, it is not possible for these values to reach 100\% without visiting every location, and the increasing stringency associated with increasing threshold values means that it should be expected that, for example, the curve for CA90 would be above the curve for CA95. The second chart is a plot of the SSE for the reconstructed map as a function of time, including the median, the inter-quartile range (IQR), as well as the 5\textsuperscript{th} and 95\textsuperscript{th} percentiles, as well as the maximum and minimum values. In a high performing system, the curves would quickly fall to near zero and would exhibit minimal spread. %

The remainder of this subsection introduces the various scenarios that were evaluated, with a brief discussion of qualitative observations. Quantitative comparative evaluation is presented in Section \ref{section_quant_eval_analysis}.

\subsubsection{Benchmark: Na\"ive Spiral Search}

\begin{figure}[t]
    \centering
    \includegraphics[width=0.75\textwidth]%
    {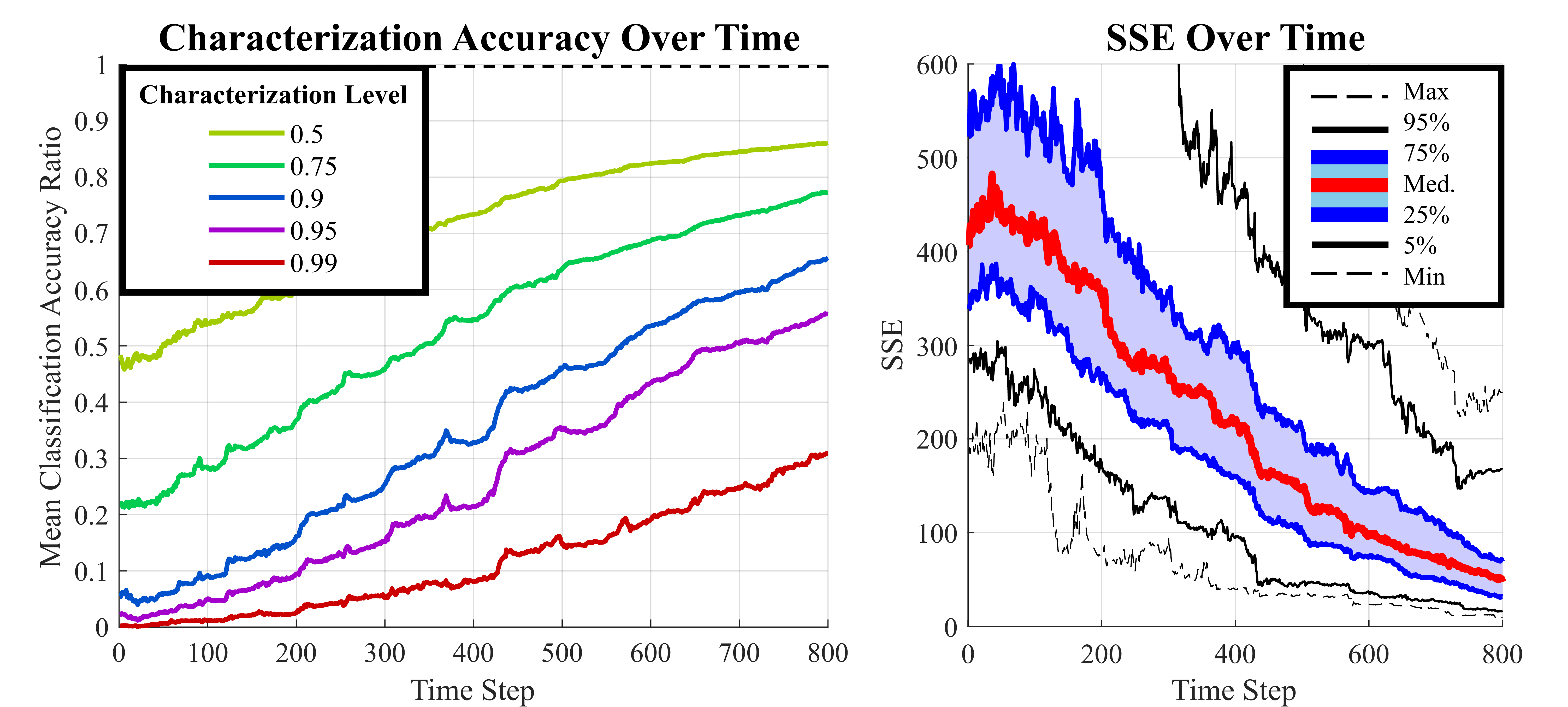}
    \caption{Characterization accuracy and SSE trends for a na\"ive spiral mapping strategy. The dashed black line represents optimal characterization accuracy of one.}
    \label{fig:result_spiral}
\end{figure}

We first consider a simple search in which a single agent begins in the center of the search domain and begins to spiral outwards. For the purposes of benchmarking we examined a spiral search pattern that completes a search of the domain in 800 time steps. Fig. \ref{fig:result_spiral} shows the results for this search pattern. When compared to the other results below, this approach struggles with higher characterization levels. Another downside to the spiral approach is that the spiral mapping strategy must commit to a specific search time at the outset, rather than having the flexibility to run the search for more or less time depending on the developing needs of the mapping mission.

\subsubsection{Benchmark: Point-To-Point search}
\begin{figure}[]
    \centering
    \includegraphics[width=0.75\textwidth]{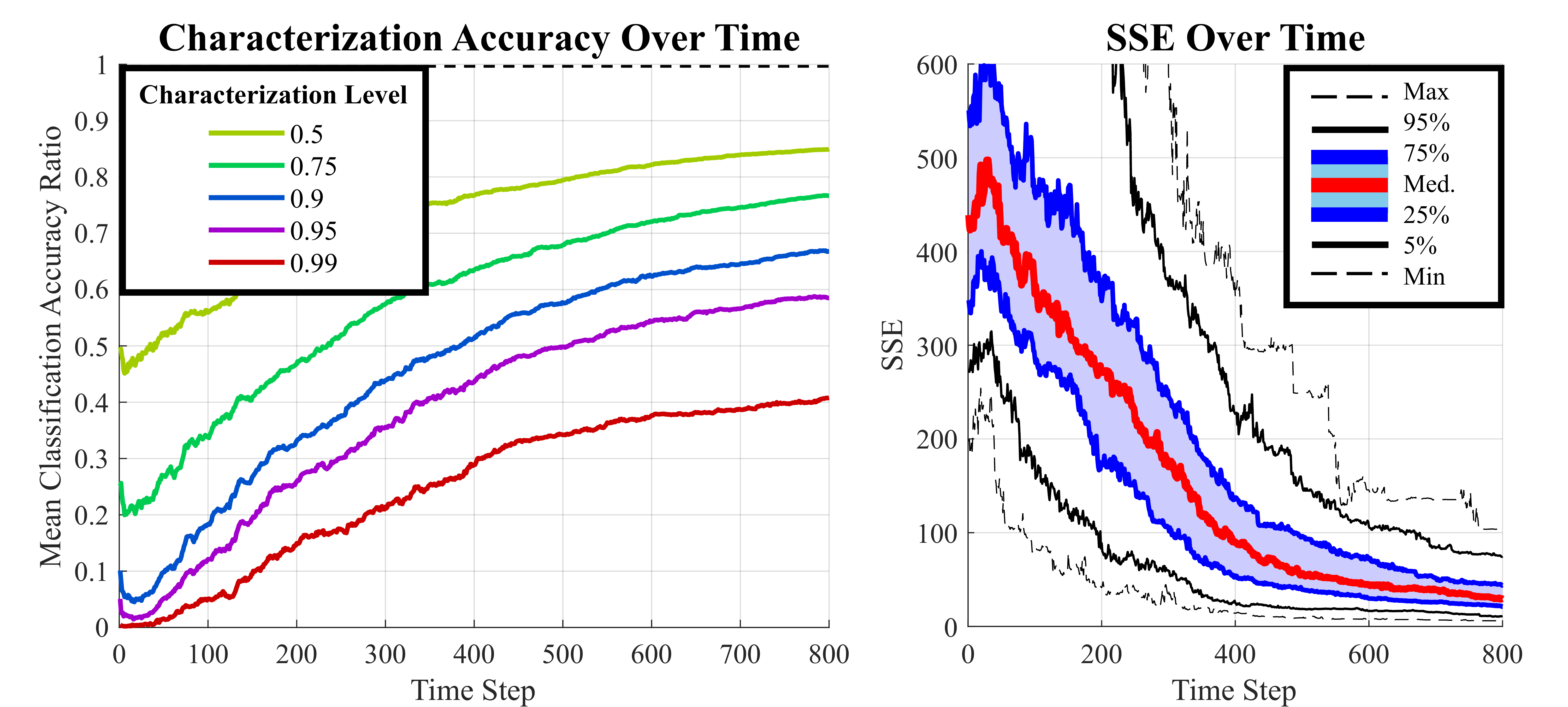}
    \caption{Characterization accuracy and SSE trends for PtP mapping strategy.}
    \label{fig:result_ptp}
\end{figure}

Next, we consider a Point-to-Point (PtP) search pattern in which a goal location is selected, and then the agent moves along the shortest path to this goal location until that goal location is reached. The process then repeats, with the agent moving from goal position to goal position, continuously sampling along the way. This strategy is similar to the approaches used in \citep{pulido2020kriging, fentanes20183}, but unlike those approaches we still assume that the robots are constantly taking measurements with each step. For goal location selection we use the same strategy as used in our full algorithm. %
Fig. \ref{fig:result_ptp} shows the results for this search strategy. In general, the improvement in characterization accuracy in the first 800 time steps follows a broadly linear trend. Similar improvement can be observed in the sum of squared errors, with the trends of the IQR exhibiting a generally-linear trend of improvement. %

We also performed simulations using a multi-agent version of the PtP strategy. The only specific new behavior added was the Voronoi partitioning approach during goal selection.

\subsubsection{SBS with one and multiple search agents}

\begin{figure}[]
    \centering
    \includegraphics[width=0.75\textwidth]%
    {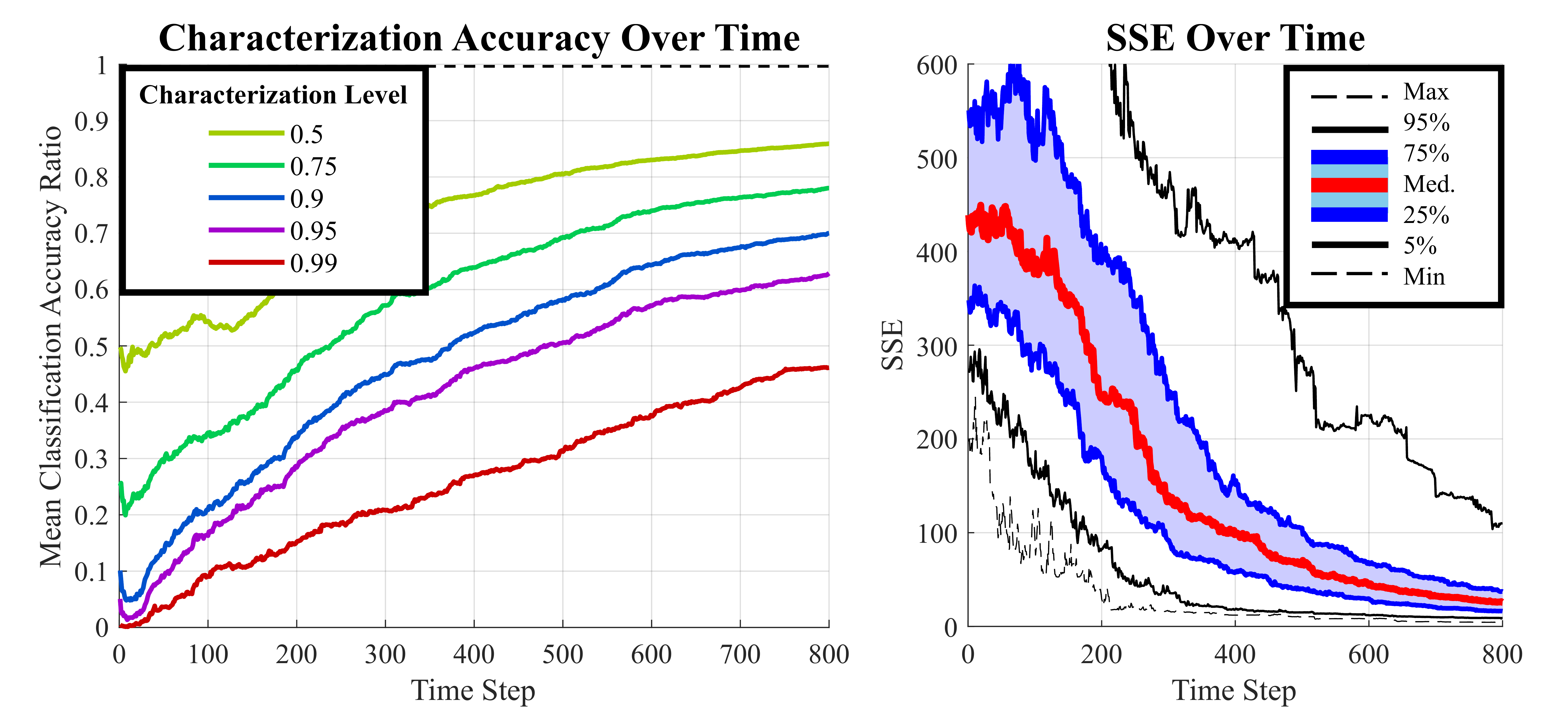}
    \caption{Characterization accuracy and SSE trends for a single agent SBS mapping strategy.}
    \label{fig:result_onerobot}
\end{figure}

We now consider the performance of SBS. Fig. \ref{fig:result_onerobot} shows the results for this search strategy for a single agent; this is shown separately from the multi-agent case to aid in comparisons to the PtP strategy. Qualitatively, the trends between the two are similar, but whereas the PtP curve for CA99 begins to plateau after T=400 (50\% of the way through the simulation), the mean curve for SBS does not show such slowdown.

\begin{figure*}[]
    \centering
    \includegraphics[width=\linewidth]{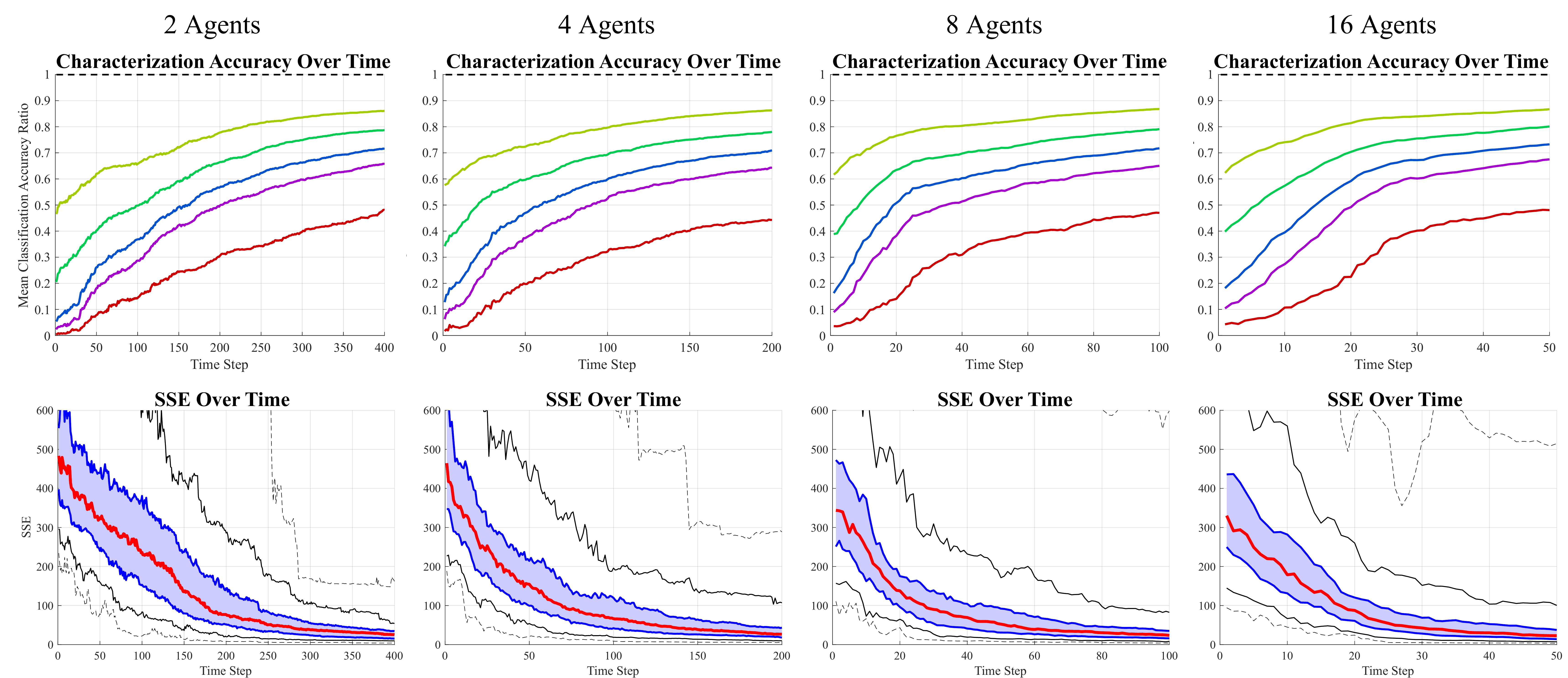}
    \caption{Characterization accuracy and SSE trends for multiple agents executing SBS. Refer to previous figures for the legend.}%
    \label{fig:result_multirobot}
\end{figure*}

Next, we examine the performance of our algorithm with multiple agents, specifically for the cases of 2, 4, 8, and 16 agents. In these experiments we adjust the number of steps each agent takes such that the total number of steps is consistent across tests. For these trials, trials with 1 and 2 agents use a value of 10 for $weight\_current\_goal$ whereas the others performed best using a value of 1 for this weight. Fig. \ref{fig:result_multirobot} compares the the results for these strategies. In general, mapping performance improved relative to the PtP for the same number of agents, and improved slightly with an increasing number of agents (quantitative discussion in Section \ref{section_quant_eval_analysis}). Notably, however, the case with 16 agents appears to perform worse than the case with 8 agents, particularly in the early stages. What this appears to reflect is some initial ``burn-in'' time for the reconstruction where the agents have not yet moved far enough to accurately reconstruct the map. Evidence to support this hypothesis can be seen in the initial upward ``swoop' seen for the CA99 curve in the cases with fewer agents. Since this sort of burn-in would take up a proportionally longer duration for the case of more agents, its effects would degrade early-search performance. Establishing the ideal swarm size relative to the search domain and timing is a problem outside the scope of this paper.

\subsubsection{Performance with Obstacles}
\begin{figure*}[htb]
    \centering
    \includegraphics[width=\linewidth]{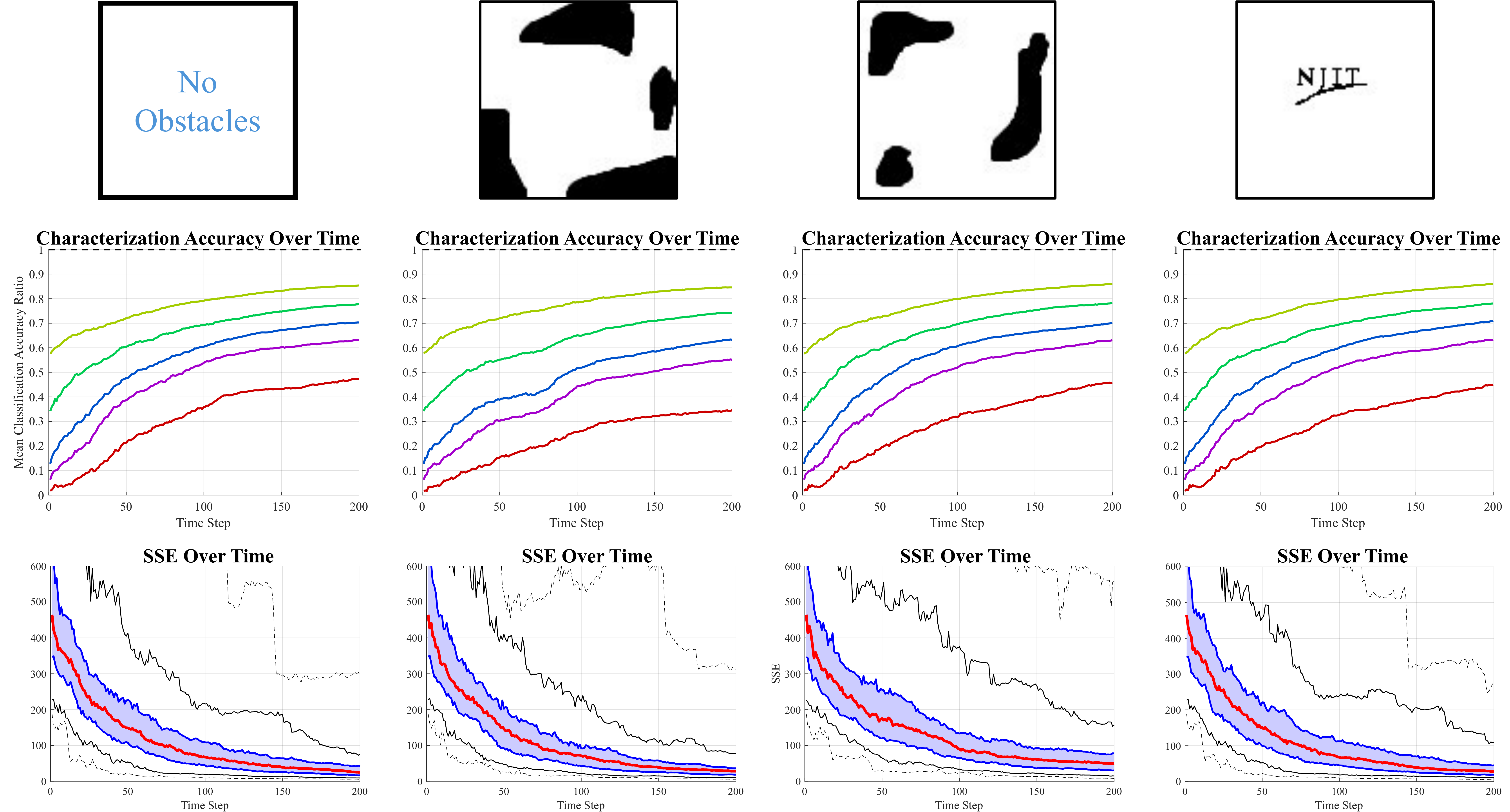}
    \caption{Characterization accuracy and SSE trends for four agents executing the SBS in the presence of different arrangements of obstacles. Performance was similar regardless of the environment.  The plot for 4 agents without obstacles is repeated from the previous figure to enable easier comparison. Refer to previous figures for the legend.}
    \label{fig:result_obstacles}
\end{figure*}

Finally, we consider the performance of the algorithm in environments where impassible obstacles are present. The ability to perform this mapping while avoiding obstacles is only possible due to the path planning that occurs as a part of our algorithm. We performed experiments for the cases of 4 agents.
Fig. \ref{fig:result_obstacles} shows both the different obstacles we used in this analysis as well as the average results for these obstacle arrangements compared against the no-obstacle results. The search performance does not appear to be greatly affected by the presence of these obstacles, minor though the first and largest set of obstacles, which reach to the edge, performed noticeably worse. When navigating these environments agents tended to traverse the outside of the obstacles, likely because the interior of these obstacles (where data was entirely censored) had high associated uncertainties, naturally drawing the agents to these locations.

\subsection{Analysis and Discussion of Simulation Results}
\label{section_quant_eval_analysis}

\begin{table*}[]
\centering
\caption{Mean and standard deviation (shown in parentheses) results of Classification Accuracy and SSE for benchmark and SBS mapping at three times.}
\label{tab:full Results}
\includegraphics[width=\textwidth]{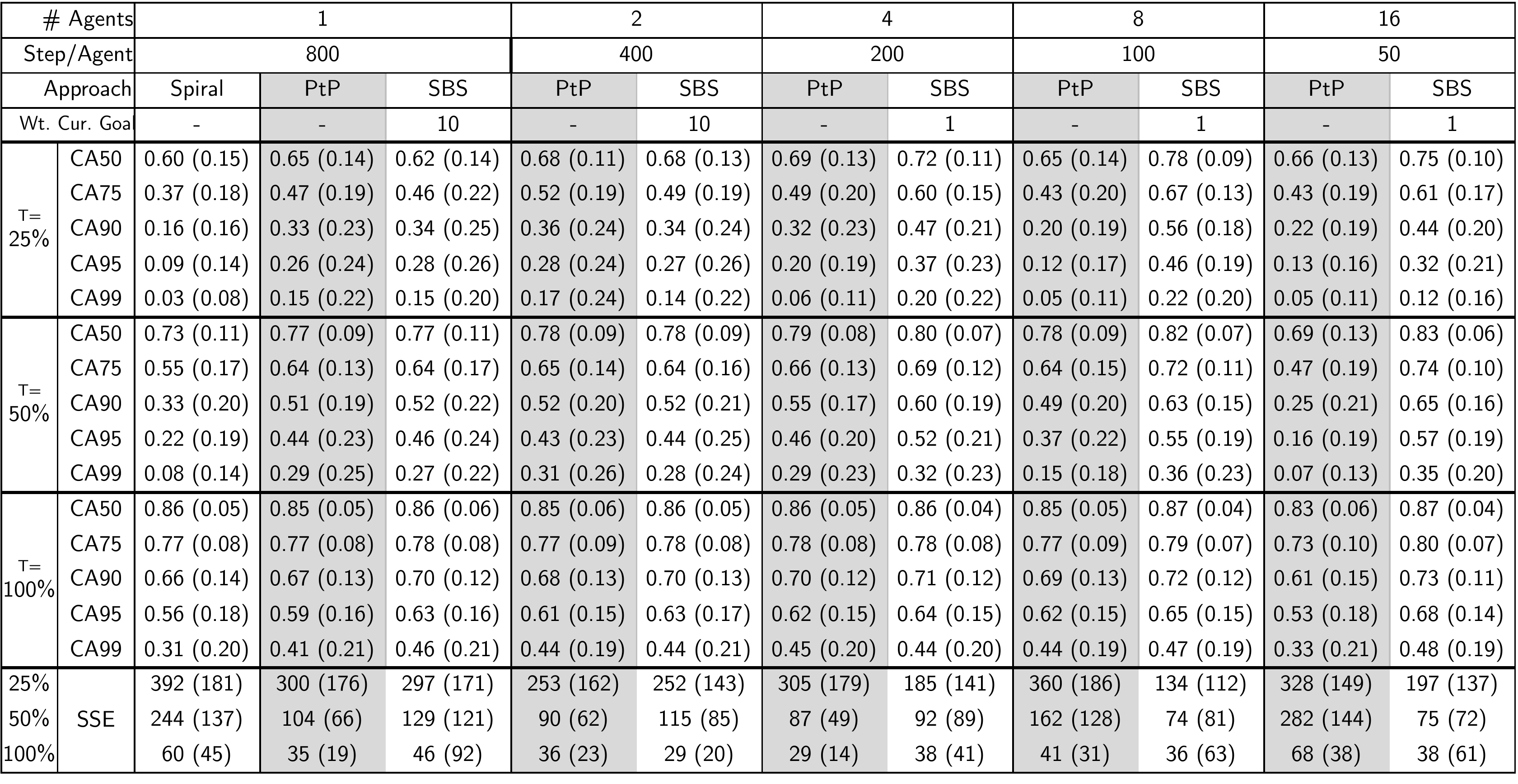}
\end{table*}

\begin{table}[]
    
\centering
\caption{P-value for alternative hypothesis of ``SBS outperforms PtP by at least 10\%.'' Cells are colored green if H\textsubscript{0} is rejected using $\alpha=0.05$ with the BH adjustment.}
\label{tab:pval}
\includegraphics[width=0.5\textwidth]%
{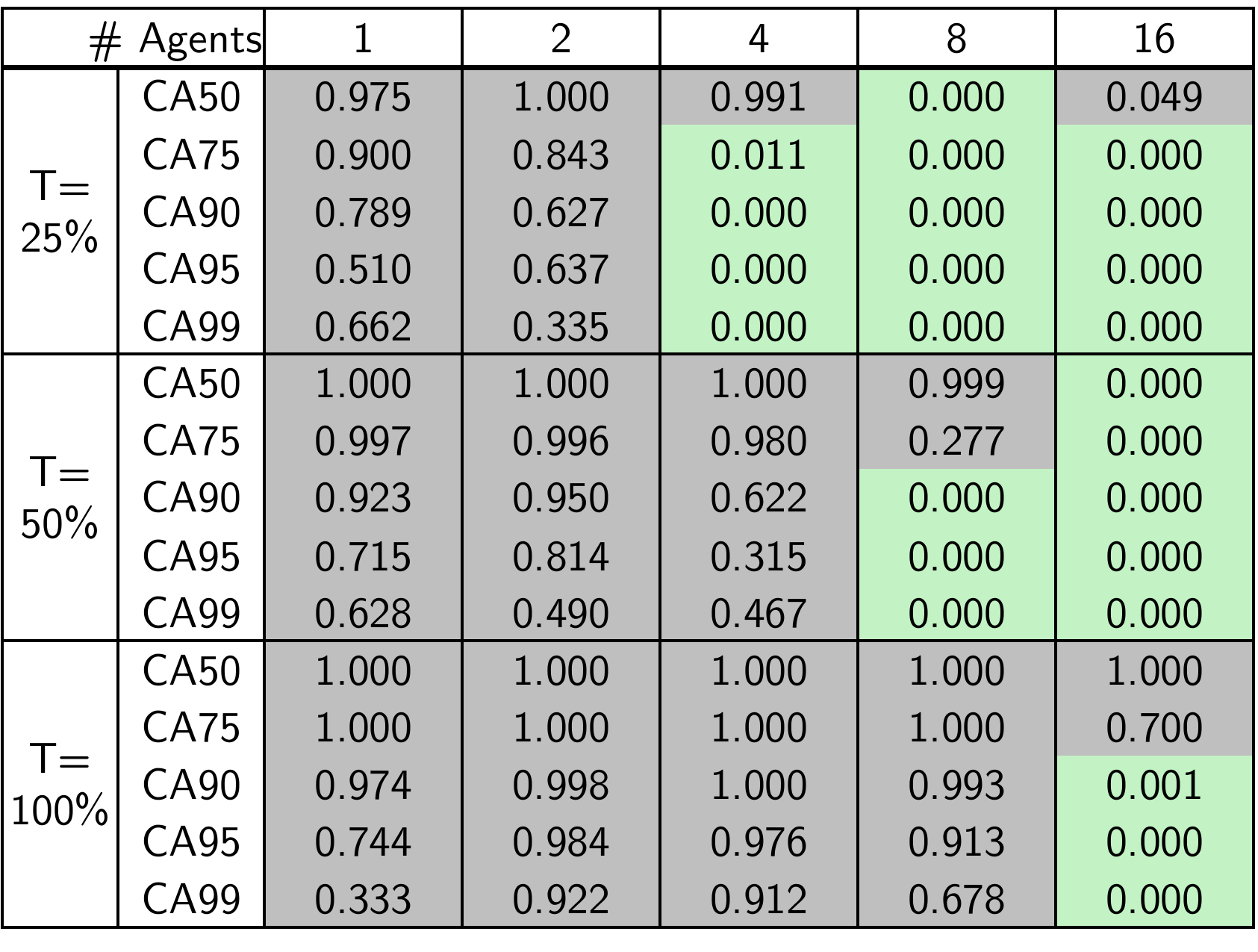}
\end{table}

\begin{table}[]
\centering
\caption{P-value for alternative hypothesis of ``SBS at N agents outperforms SBS at $N/2$ agents by at least 5\%.'' Cells are colored green if H\textsubscript{0} is rejected using $\alpha=0.05$ with the BH adjustment.}
\label{tab:better_than_lower}
\includegraphics[width=0.5\textwidth]%
{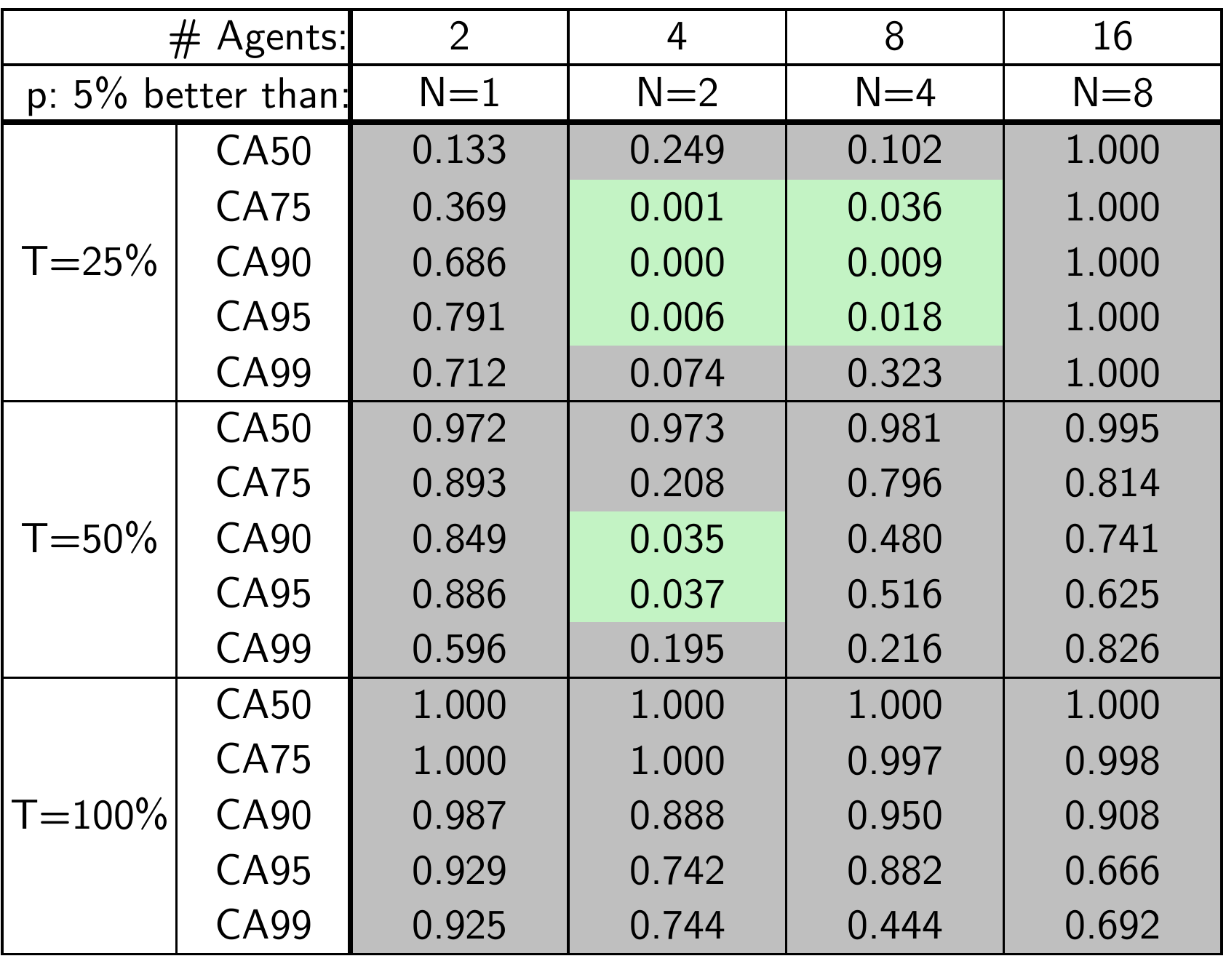}
\end{table}

The quantitative results of these experiments are summarized in Tables \ref{tab:full Results}. For each configuration tested, the mean and standard deviation are reported for 100 simulated trials each, and these results are given at three temporal milestones (25\%, 50\%, and 100\% of the total 800-step search budget). Along with the knowledge that each number results from 100 trials, this table contains sufficient information to perform the other statistical tests presented in this section, but results are rounded for readability. 

As a first-glance analysis in comparing the means values at each timestep and agent count, the general trend is observed that SBS consistently demonstrates superior performance to the Na\"ive spiral and PtP benchmarks, particularly at earlier temporal milestones and higher characterization thresholds: while all methods performed similarly at CA50, especially at T=100\%, means for CA99 at N = 4 and 8 for the earlier temporal milestones can outperform PtP by two to four times. This reflects a distinct advantage in quickly detecting the most critical regions. 

This advantage is somewhat undercut by the relatively high standard deviations for all tests, which are consistent across approach used and the number of agents. What this likely reflects is the inherent unpredictability and ``roughness'' of the Fractional Brownian Fields used for the maps. These results therefore reflect the performance of each approach in a particularly challenging scenario. What this means in practice is that the standard deviations presented are significantly influenced by the underlying variation of the map itself, rather than being a pure reflection of the mapping behavior itself. Therefore, tests for statistical significance must overcome the presence of significant random noise.

The first statistical test we perform is a 2-sample T-test to determine if, in aggregate, the SSE of SBS outperforms PtP. This is performed by taking the mean of means and mean of variances for each, making each population represent 500 samples each. For these T-tests, H\textsubscript{0} is that the mean of SBS population is less than 10\% better than the mean of the PtP population. Using $\alpha=0.05$, we reject H\textsubscript{0} for T=25\% ($p=3\times10^{-11}$) and T=50\% ($p=10^{-8})$, but we fail to reject H\textsubscript{0} for T=100\% ($p=0.51$). What this reflects is an overall advantage for SBS earlier in the mapping process, though the exact advantage depends on the robot count. 

Next, we investigate whether SBS outperforms PtP at each of the temporal milestones and characterization thresholds. Arbitrarily, we use a desired an improvement of at least 10\%. H\textsubscript{0} is therefore that SBS is less than 10\% better than PtP. We apply 2-sample T-tests for each permutation of time, characterization, and agent count, with $\alpha=0.05$. Due to the large number of tests, we apply the Benjamini-Hochberg (BH) procedure \cite{benjamini1995controlling} to reduce the risk of false positives. The results of this analysis are shown in Table \ref{tab:pval}, with green highlights indicated instances where we reject H\textsubscript{0} and conclude that the performance of SBS is at least 10\% better. Checking the reverse analysis indicated that PtP was never statistically significantly better than SBS. From this, we again see the trend that SBS performs better at earlier temporal milestones and at higher agent counts.

Finally, we seek to establish the degree to which higher agent counts benefit mapping quality. To do this, we use a similar approach to before, but this time we examine whether the characterization accuracy with N agents is at least 5\% better than the characterization accuracy with N/2 agents. The results of this analysis are shown in Table \ref{tab:better_than_lower}. In this case, the relatively large standard deviations resulted in a less clear advantage, though advantages to an increased number of search agents again appeared in early temporal milestones. Importantly, flipping the analysis (i.e., examining whether fewer agents outperformed more agents) failed to reject H\textsubscript{0} in all instances except for T=25\% for the case with 16 agents, as discussed earlier. 

From these statistical analyses we can assert that SBS exhibits significant advantages against the PtP benchmarks. Overall, SBS demonstrates statistically significant improvements over PtP at early time steps and higher characterization thresholds, while differences diminish at later stages.

%% file: experimental_results.tex
\label{sec:experiments}
\label{Section_experiments}

Having demonstrated the algorithm in simulation, we now turn our attention to implementing the algorithm using the Bionic Swarm platform. The goal also is to demonstrate the ability to deploy the Bionic Swarm platform to execute a swarm algorithm outdoors. Notably, Our goal in this section is not to match the statistical validation and scientific reproducibility of the simulation results, but rather to understand the general performance of the algorithm in a real-world setting and to understand what real-world factors could be considered towards algorithm refinement. This is consistent with norms in the literature where robust simulation analysis is paired with a much smaller number of in-lab experiments.

\subsection{Study Site}
\begin{figure}
    \centering
    \includegraphics[width=0.75\textwidth]%
{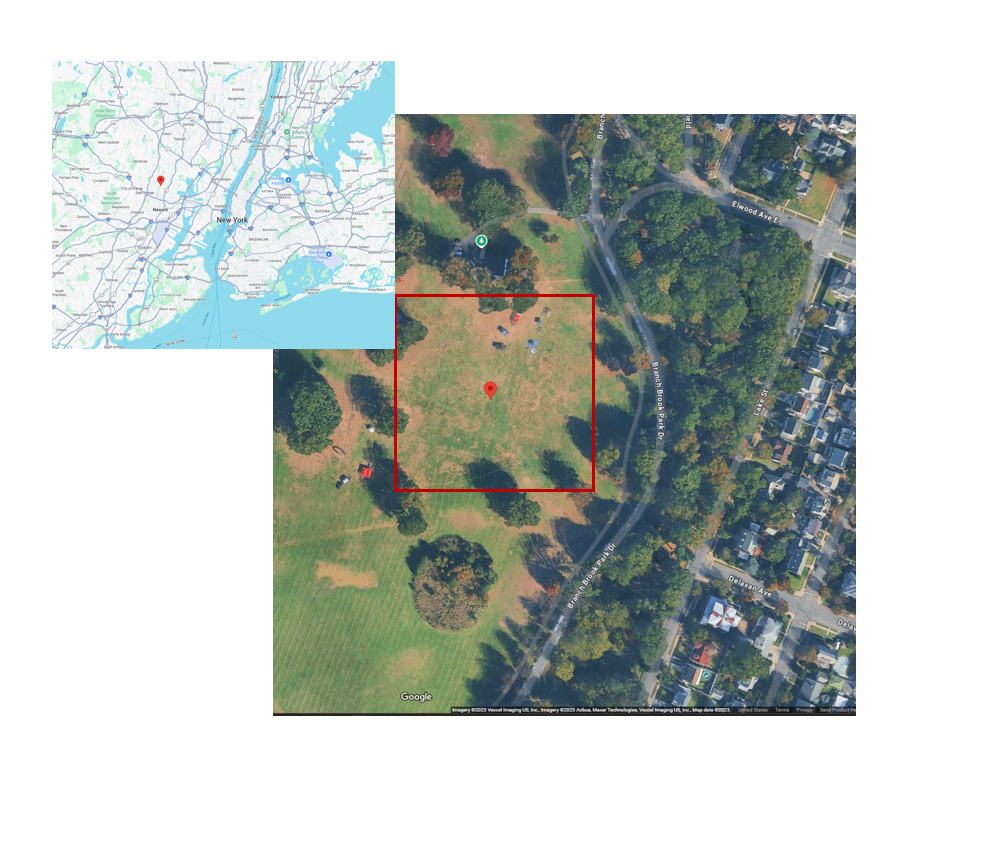}
    \caption{Location and layout of data collection site: Branch Brook Park, Newark, New Jersey ($150 \times 150$ m test area).}
    \label{fig:dataset-location}
\end{figure}

\begin{figure}
    \centering
    \includegraphics[width=0.4\textwidth]%
{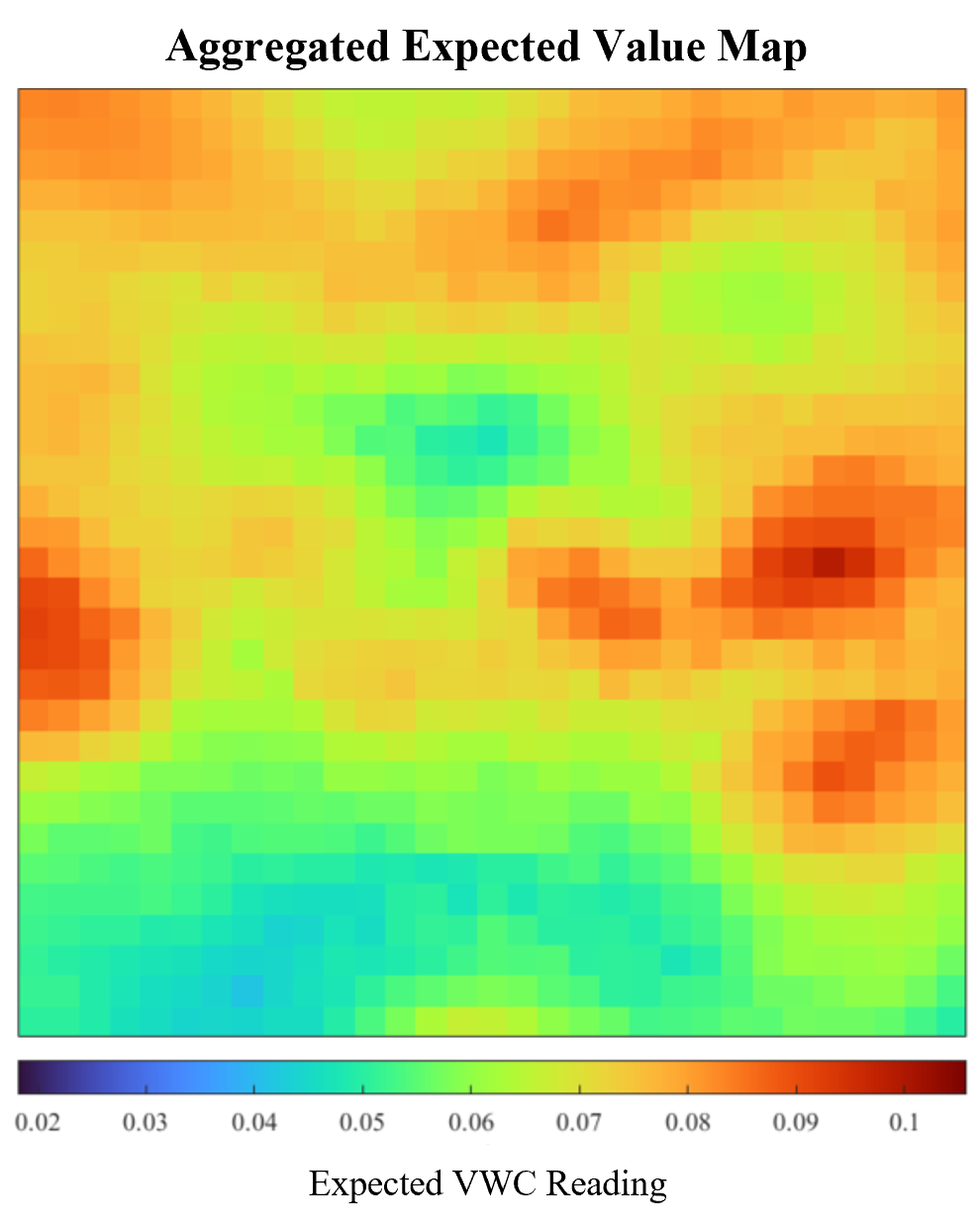}
    \caption{Ground-Truth Map of Volumetric Water Content (VWC) Constructed from Aggregated Field Measurements via Ordinary Kriging. The region represents the highlighted square from Fig. \ref{fig:dataset-location}}.
    \label{fig:true-map}
\end{figure}

Fig.~\ref{fig:dataset-location} shows the location of the experimental site: Branch Brook Park in Newark, New Jersey. The selected test site provided a continuous 150~m $\times$ 150~m meter soil patch without interruptions such as asphalt, concrete, or other artificial structures. This ensured that observed heterogeneity could be attributed solely to soil variation. We generated the ground-truth map shown in Fig.~\ref{fig:true-map} by aggregating all datasets collected across multiple runs in the same field and reconstructing the map using Ordinary Kriging.

\begin{figure}
    \centering
    \includegraphics[width=0.4\textwidth]%
{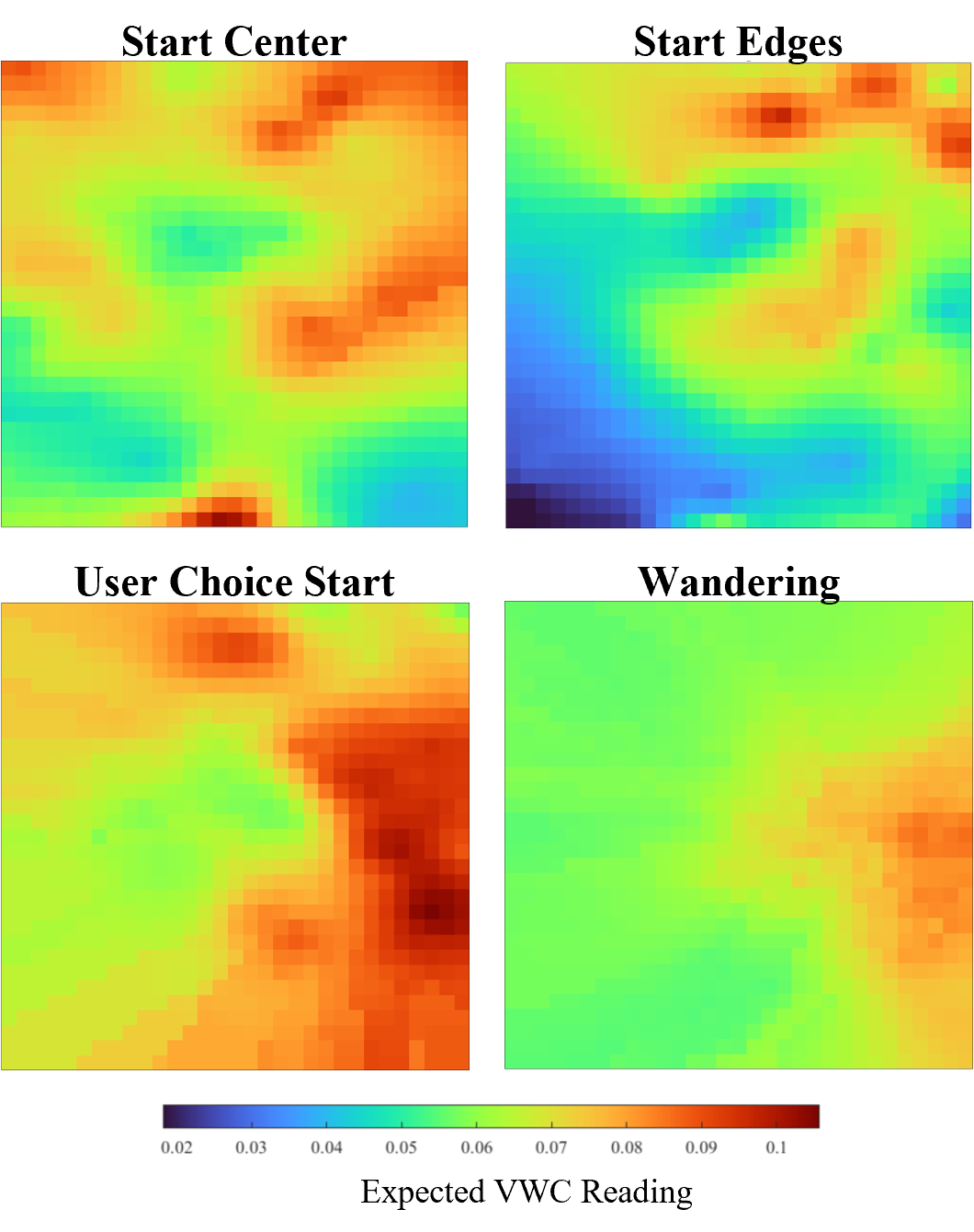}
    \caption{Comparing different starting strategies as well as an uninformed wandering search after 80 readings.}
    \label{fig:comparestart}
\end{figure}

\subsection{Experimental Procedure}
Four operators simultaneously collected data within the 150~m $\times$ 150~m field. The first device was initialized at the center, defining the reference grid. The starting positions of the other three operators varied depending on the test variation: in some runs they began near the edges, in others close to the center, and in others at random positions. The experiment was conducted using grid cells of 10 m $\times$ 10 m, based on preliminary tests which showed a worst-case $\pm$ 5 m variation for GPS accuracy near campus buildings.

Although the system architecture is not inherently limited in the number of participants and can be readily scaled to larger groups, four was selected as the number of human agents to match what was done in simulation. Limited preliminary trials were also conducted with up to six operators, showing consistent system behavior.

Ordinary Kriging with a spherical variogram model was applied after the first three measurements, selected because a bounded, finite-range model aligns with field-scale soil heterogeneity and has shown strong empirical performance in recent soil-moisture mapping studies \cite{Brown2023VZJ,Mensah2023CJSS}.
After each measurement, the server assimilates all sensor readings up to that moment and performs Ordinary Kriging to estimate expected values and uncertainties at every grid cell. These estimates are fed into the SBS algorithm and combined with other factors via specified weights to determine the direction that the users should move in order to take their next measurement. Note that one major difference in the real-world experiments is that agent motion is no longer synchronized since users move at their own pace, highlighting the sort of assumption that can be easy to overlook when developing simulations.

We tested four approaches to mapping, two of which remove agency from the human users, and two which give the users more influence over the course of the experiment:
\begin{enumerate}
    \item SBS exploration starting from the center of the field.
    \item SBS exploration starting from the edge of the field.
    \item SBS exploration starting from user-selected starting positions.
    \item  Uninformed ``Wandering'' strategy without following SBS.
\end{enumerate}
The aim was twofold: (1) Determine whether following the score-based algorithm improves accuracy and efficiency in detecting regions of high volumetric water content (VWC) compared to unguided wandering; and (2) investigate how the choice of initial position influences map accuracy and convergence speed. 

\subsection{Experimental Results}
As shown in Fig.~\ref{fig:comparestart}, the selection of starting locations is important to how well the system is able to reconstruct the map. After 80 readings, the
\emph{Start Center} and \emph{Start Edges} variants capture the overall shape of the true map
gradient with good fidelity (especially for the \emph{Start Center} condition), while \emph{User-Selected Start Positions} captures some features of the true map but entirely misses others, likely due to the biases of the users when choosing a start location. 
\emph{Wandering}
yields a diffuse, low-contrast map that underestimates the peak VWC and departs
visibly from the ground truth.

\begin{figure*}
    \centering
    \includegraphics[width=\linewidth]{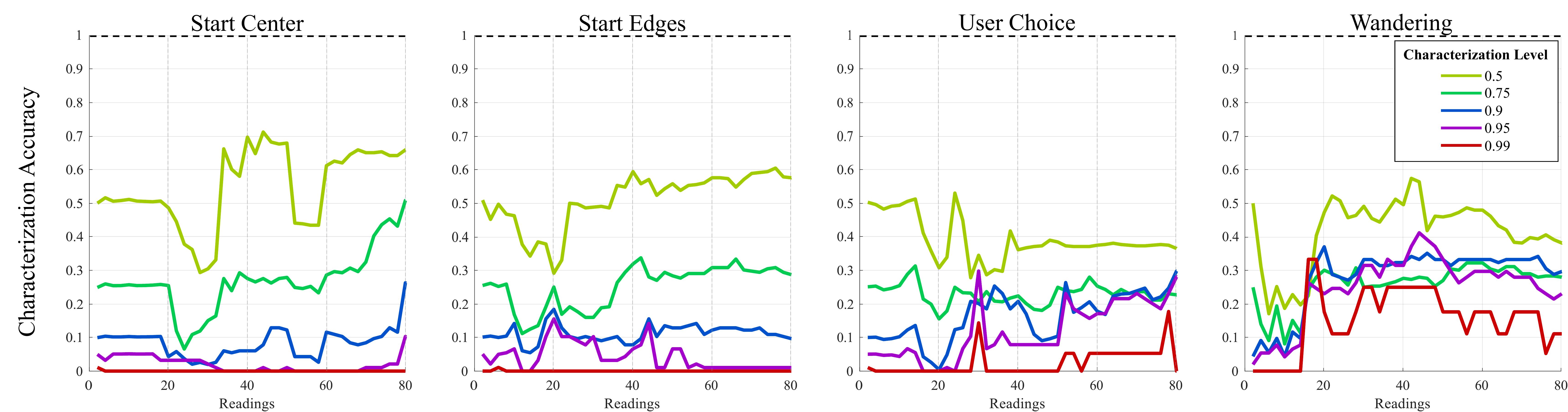}
    \caption{Comparison of characterization accuracies for the four strategies examined in the experiments.}
    \label{fig:compare_characterization}
\end{figure*}

Fig. \ref{fig:compare_characterization} shows a quantitative comparison of these experimental results using the characterization accuracy metric. A key difference relative to the graphs from previous sections is that the X axis represents the measurement number, rather than the actual timing of the measurement, since people would naturally walk at different rates.
While each graph only represents a single run (in contrast to the 100 runs used to generate the curves shown in the simulation results section), the curves for \emph{Start Center} and \emph{Start Edges} broadly resemble the gradual improvement then plateauing of the characterization accuracy seen in the simulated results. In the case of \emph{User Choice Start} and \emph{Wandering}, however, user biases confounded the overall process, highlighting potential risks for integrating human users so deeply into the operation of the system.

We also performed several trials with different cell sizes to understand what, if any, trade-offs there would be to forcing more or less frequent sampling. Small cells (5 m $\times$ 5 m) produced the higher precision reconstructions, particularly in identifying localized regions of elevated moisture, but required greater measurement effort. Conversely large cells (20 m $\times$ 20 m) enabled rapid coverage but smoothed out finer details. These results are in-line with intuitive expectations.

%% file: discussion.tex
\begin{table*}[]
\centering
\caption{Comparison of deploying traditional robot swarms to the Bionic Swarm.}
\label{tab:discussion_compare}
\begin{tabularx}{\textwidth}{|ll|X|X|}
\hline
\cellcolor[HTML]{9B9B9B}\textit{}    & \cellcolor[HTML]{9B9B9B} & \textbf{Traditional Robotic Swarm}   & \textbf{Bionic Swarm}                                     \\ \hline
                                     & Upfront Cost      & \textbf{High} (can be \$100k+)                  & \textbf{Low} (test-specific sensors)                      \\ \cline{2-4} 
\multirow{-2}{*}{\textit{Financial}} & Operational Cost         & \textbf{Moderate} (maintenance)               & \textbf{Very Low}                                         \\ \hline
                                     & Development              & \textbf{Very High} (months to years)          & \textbf{Low} (about one month)                            \\ \cline{2-4} 
                                     & Deployment               & \textbf{Moderate} (transport and calibration) & \textbf{Very Low} \newline (open web app and connect to Bluetooth) \\ \cline{2-4} 
\multirow{-3}{*}{\textit{Time}}      & Approvals                & Site permits                                 & IRB, depending on the experiment                          \\ \hline
                                     & Operator Count           & 1-2 highly skilled operators                 & 1 supervisor \newline + minimally-trained operators                \\ \cline{2-4} 
\multirow{-2}{*}{\textit{Personnel}} & Training Required        & \textbf{High}                                 & \textbf{Minimal}                                          \\ \hline
\end{tabularx}
\end{table*}

Experimental results are broadly consistent with the simulated results: the Bionic Swarm system was able to reconstruct the spatial distribution of soil moisture with reasonable fidelity, and the evolution of characterization accuracy over time followed patterns consistent with those predicted in simulation. As expected, performance degraded when greater autonomy was given to human operators, as seen in the user-selected starting positions and uninformed wandering scenarios. This outcome highlights a central insight of the work: algorithmic guidance can significantly improve the efficiency and reliability of distributed sampling tasks, even when executed by human agents rather than autonomous robots.

Differences such as worse performance for the highest classification levels could be due to a number of factors. The first is of course that the experimental results represent single runs of the system whereas the simulation results are each represented by 100 runs of the system. The second is that we do not have access to the true map, instead relying on a map reconstructed from several runs throughout the day, during which time the underlying distribution of soil moisture is likely to have changed in some ways due to evaporation or other phenomena, resulting in a comparative baseline that does not exactly match the conditions during a specific experiment. Finally, this could represent a difference in performance for the system when the underlying distribution is a Fractional Brownian Field (as with the simulations) versus the real-world distribution. It is important to note, however, that these are problems inherent to any field experiment.

Beyond validating the Score-Biased Search (SBS) algorithm, the primary contribution of this work lies in demonstrating a new paradigm for shrinking the gap between simulation and real-world testing. %
To use this particular research as a case study, when initially considering how to perform field tests we had explored purchasing four Husky Jackal UGV robots \cite{jackalUGV2025}, and then deploying the system at a similar location as the one used for the experiments presented in this paper. The expected time to program the robots, obtain approvals for reserving exclusive use of the test area, and performing of the experiments was approximately one year. Combining the associated labor costs with the cost of the robots themselves, such a project would have easily had a total cost in excess of \$100,000. This high cost made it very difficult to justify any sort of real-world validation for the presented mapping strategy. 

In stark contrast to this was the amount of effort required to perform the experiments presented in the Section \ref{sec:experiments}. Instead of spending a year to ensure that  field-deployed robots would operate safely and effectively, human users are already skilled at navigating open fields. Further, because people walking around a public park is both common and expected, it was unnecessary to reserve exclusive use of the park, though it did require verifying that the experiments were IRB exempt. In total, preparation for the experiments beyond the development of the generic Bionic Swarm system required only one month of part-time work to translate code from the simulations to function on the server. Likewise, the Bionic Swarm system drives significant hardware savings: whereas robot hardware would have cost several tens of thousands of dollars, the price for assistance from undergraduate students is generally accepted to be about one slice of pizza.

The limitations of this approach must be considered, however. One concern is reproducibility, since the use of human operators introduces a potential source of significant experiment-to-experiment variance that would not necessarily manifest when using real robots. The approach is therefore not well suited to examining the effects of subtle variations in algorithm parameters.

Algorithm conformance and execution accuracy are other sources of potential concern: even if operating with full intent to exactly follow instructions, the human operators could introduce their biases and preferences into their task. While a direct experimental comparison to a robotic swarm is beyond the scope of this study, prior robotic mapping systems (e.g., \cite{fentanes20183, pulido2020kriging}) typically assume synchronized motion and more precise positioning, whereas the bionic swarm introduces asynchronous motion and human decision variability. For example, in a mapping task such as the one presented here, the system could command a person to walk through a puddle, which the person would, of course, not want to do, preventing the algorithm from being tested as-implemented. However, this addition of human preference could be useful in refining and adapting the algorithm for robot use: the roboticists may not have previously considered the possibility of an errant puddle. In such a case, a similar on-the-fly modification would need to be made, resulting in the same failed experiment, one that would have required significantly more planning and investment.

The system as-presented is also only best suited to trialing high-level logic and sensor integration: research questions that rely on high-precision or high-speed actuators are unlikely to involve actions that are easily performed by a human operator. However, because the system integrates with a standard microcontroller, it would be possible to add actuators or similar hardware to the system to extend the capability of the system to match operational needs, though the system is likely not well suited to testing hypotheses regarding robot dynamics or task execution speed. 

Finally, whereas robots can be commanded to work for months at a time, there are practical limits to what can be asked of human volunteers and student workers, meaning that it becomes difficult to run enough trials to gain statistical insights into the performance of the system. This human time-availability limitation also impacts the ability to deploy the large-scale swarms that are possible in simulation

Taken together, these findings position the Bionic Swarm as a complementary tool within the broader workflow of swarm robotics research. It enables low-cost, rapid, and accessible real-world validation at early stages of development, helping to de-risk subsequent investments in full robotic systems. Rather than replacing simulation or robotic experimentation, the Bionic Swarm fills a critical gap between them, accelerating iteration cycles and expanding access to real-world testing.

%% file: conclusions.tex
The Bionic Swarm system enabled us to quickly and inexpensively validate the function of the SBS algorithm. This was made possible by breaking free of the paradigm of using robots to validate robot-focused algorithms and instead abstracting away many of the difficult-to-implement specifics of robot operation to human users. This system is not specific to the algorithm and experiments presented here, but is generic and could be used in a wide variety of contexts with minimal upfront work. Although robot soil mapping, as well as unguided human sampling has been explored in earlier work, the Bionic Swarm is the first system to our knowledge that closes the loop, using real-time spatial models to actively guide a swarm of human agents towards efficient environmental mapping. It represents a new paradigm for rapid, low-cost, real-world validation for swarm algorithms.

While fully-robotic experiments will remain the gold standard in algorithm validation, using a bionic approach provides new avenues for earlier real-world testing, as well as for testing ideas that may not have previously warranted investing the time and effort to try in the real world. This makes real-world testing accessible for more research groups while also enabling well-funded research groups the opportunity to quickly iterate on sensors and algorithms so that their eventual full-scale tests are of higher quality. This is of particular benefit to field and swarm roboticists, since these fields tend to have high  experiment costs and have a high standard for system reliability. 

Future development on the Bionic Swarm platform will focus on improving the ease of applying the system towards new applications and incorporating new types of sensors. At the same time, we will work towards backend improvements to increase system scalability and efficiency. While experiments were conducted with four operators (and preliminary tests with six), the reduced swarm size relative to simulation (up to 16 agents) represents a key limitation. Future work will evaluate scaling effects using larger human cohorts or hybrid setups. We will also explore opportunities to use the system towards applying existing swarm algorithms in real-world environments and to identify ways to incorporate swarm approaches to existing applications. We additionally wish to eventually test the bionic swarm approach against a true robot swarm, both to validate that the systems perform similarly as well as to understand the benefits, drawbacks, and opportunities of adding a human factor in the process of algorithm development and validation.